\documentclass[letterpaper, 10 pt, conference]{icra2024}  

\IEEEoverridecommandlockouts                              

\usepackage{amsmath,amsfonts}
\usepackage{algorithm}
\usepackage{array}
\usepackage[caption=false,font=normalsize,labelfont=sf,textfont=sf]{subfig}
\usepackage{textcomp}
\usepackage{stfloats}
\usepackage{url}
\usepackage{verbatim}
\usepackage{graphicx}
\usepackage{cite}
\usepackage{times}
\usepackage{epsfig}
\usepackage{amsmath}
\usepackage{amssymb}
\usepackage{multirow}
\usepackage[table]{xcolor}
\usepackage{algpseudocode}

\algrenewcommand\textproc{}

\title{\LARGE \bf Filling Missing Values Matters for Range Image-Based Point Cloud Segmentation
}

\author{Bike Chen, Chen Gong, and Juha Röning
}

\begin{document}

\maketitle
\thispagestyle{empty}
\pagestyle{empty}

\begin{abstract}
Point cloud segmentation (PCS) plays an essential role in robot perception and navigation tasks. To efficiently understand large-scale outdoor point clouds, their range image representation is commonly adopted. This image-like representation is compact and structured, making range image-based PCS models practical. However, undesirable missing values in the range images damage the shapes and patterns of objects. This problem creates difficulty for the models in learning coherent and complete geometric information from the objects. Consequently, the PCS models only achieve inferior performance. Delving deeply into this issue, we find that the use of unreasonable projection approaches and deskewing scans mainly leads to unwanted missing values in the range images. Besides, almost all previous works fail to consider filling in the unexpected missing values in the PCS task. To alleviate this problem, we first propose a new projection method, namely scan unfolding++ (SU++), to avoid massive missing values in the generated range images. Then, we introduce a simple yet effective approach, namely range-dependent $K$-nearest neighbor interpolation ($K$NNI), to further fill in missing values. Finally, we introduce the Filling Missing Values Network (FMVNet) and Fast FMVNet. Extensive experimental results on SemanticKITTI, SemanticPOSS, and nuScenes datasets demonstrate that by employing the proposed SU++ and $K$NNI, existing range image-based PCS models consistently achieve better performance than the baseline models. Besides, both FMVNet and Fast FMVNet achieve state-of-the-art performance in terms of the speed-accuracy trade-off. The proposed methods can be applied to other range image-based tasks and practical applications.

\end{abstract}

\section{INTRODUCTION}
The purpose of point cloud segmentation (PCS) is to assign each point a label. The task plays an important role in robot perception~\cite{joint3dinstance_zhou2020} and navigation~\cite{semantic_slam_2019} tasks because the segmentation results on light detection and ranging (LiDAR) data help robots gain a direct understanding of their physical environments.

To efficiently parse large-scale outdoor point clouds~\cite{semantickitti_2019_behley,semanticposs_2020,nuscenes_panoptic}, the range image representation of the data is commonly adopted. This image-like representation makes unordered, sparse, irregular, and large-scale points in a scan compact and structured. Built on the generated range images, corresponding models~\cite{squeezesegv2,squeezesegv3_2020,fidnet_2021,rangenet++,cenet_2022,rangevit_2023,rangeformer_2023} are usually efficient and practical, because they do not require high computational cost when compared with point-based approaches~\cite{randla_2020,pointnet++_2017} and voxel-based methods~\cite{cylindrical3d2021,spvnas_2020,spherical_transformer_2023}.

\begin{figure}[t]
	\centering
	\includegraphics[width=0.91\columnwidth]{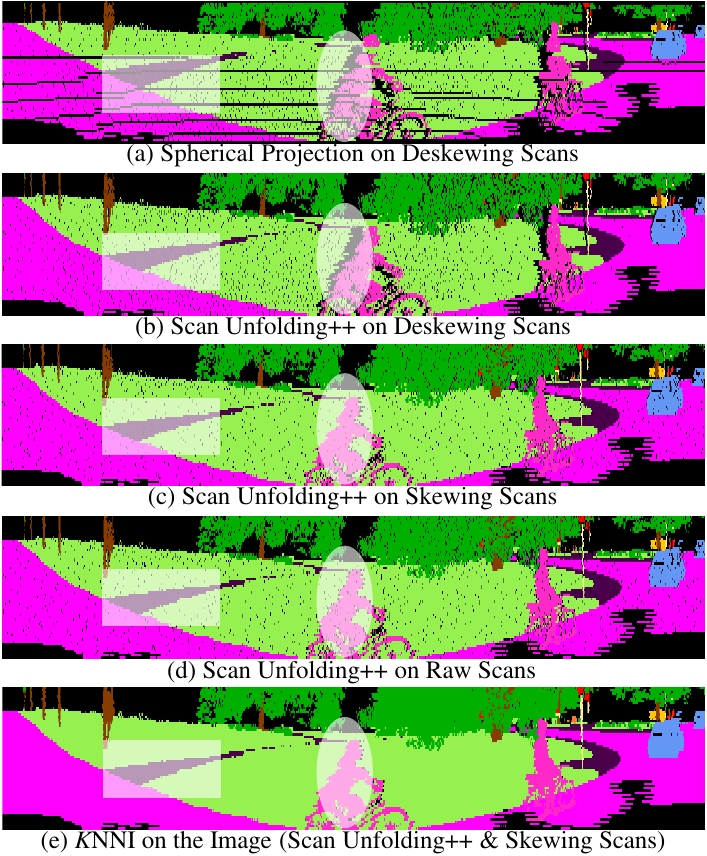}
	\caption{(a) The image produced by spherical projection~\cite{rangenet++} on the deskewing scan in the SemanticKITTI~\cite{semantickitti_2019_behley} dataset. There are many missing values. Specifically, lasers not evenly spaced in the vertical direction lead to black horizontal lines (emphasized by the white rectangle mask). Besides, deskewing scans (after motion compensation) cause large missing values (highlighted by the white ellipse mask). (b) The image generated by the proposed scan unfolding++ on the deskewing scan. All black horizontal lines (missing values) have been removed. (c) The image made by scan unfolding++ on the skewing scan. The large missing values within the white ellipse mask have been filled in. (d) The image produced by scan unfolding++ on the raw scan. It is used for comparison with the image in (c). (e) The image after applying the proposed range-dependent $K$-nearest neighbor interpolation ($K$NNI) on the image in (c). Many missing values (small black points) have been filled in valid values. All objects such as the \textit{bicyclist}, \textit{car}, and \textit{road} appear coherent and complete.}
	\label{fig:sp_su_deskew_skew_raw_knni}  
\end{figure}

However, when training PCS models on the prepared range images, we find that missing values in the range images degenerate the performance of PCS models. Three factors cause the missing values: (1) The unreasonable projection approach, namely spherical projection~\cite{rangenet++}, causes scan lines to overlap, especially when the lasers~\cite{hdl_64e_s2_manual} in the vertical direction are not evenly spaced (see black lines in Fig.~\ref{fig:sp_su_deskew_skew_raw_knni}(a)). (2) The deskewing scans~\cite{semantickitti_2019_behley} (\textit{i.e.}, after motion compensation) lead to the missing values in the horizontal direction in the range images (see the black holes emphasized by the white ellipse mask in Fig.~\ref{fig:sp_su_deskew_skew_raw_knni}(a)). (3) The inherent properties of the LiDAR sensor~\cite{hdl_64e_s2_manual} result in the missing values (see many small black pixels in Fig.~\ref{fig:sp_su_deskew_skew_raw_knni}). For example, certain lasers fail to receive valid photons as their laser beams fly too far to be received. And some laser beams are absorbed by absorbing materials. 

The missing values inevitably bring difficulties in training PCS models to achieve optimal performance. Specifically, (1) the missing values damage the shapes and patterns of objects in the range images, thereby challenging the PCS models to effectively learn coherent and complete geometric information from incoherent and incomplete objects (see the broken shape of the bicyclist in Fig.~\ref{fig:sp_su_deskew_skew_raw_knni}(a)). (2) The undesirable missing values expect that the models should possess an additional ability to predict them. This can distract attention from recognizing valid values.

In addition, almost all existing range image-based models fail to consider the missing values and exhibit inferior performance. Specifically, most models~\cite{squeezeseg,squeezesegv2,rangenet++,squeezesegv3_2020,fidnet_2021,cenet_2022,rangevit_2023,rangeformer_2023} adopt spherical projection to prepare range images. The work~\cite{scan_based_projection} utilizes scan unfolding to generate range images, but the proposed algorithm can only be applied to the raw LiDAR data. Besides, none of these models consider the negative impact of the deskewing scans or the properties of the LiDAR sensor. Therefore, the models' performance is suboptimal due to the challenge of learning features from incoherent and incomplete objects.

\begin{figure*}[t]
	\centering
	\includegraphics[width=1.5\columnwidth]{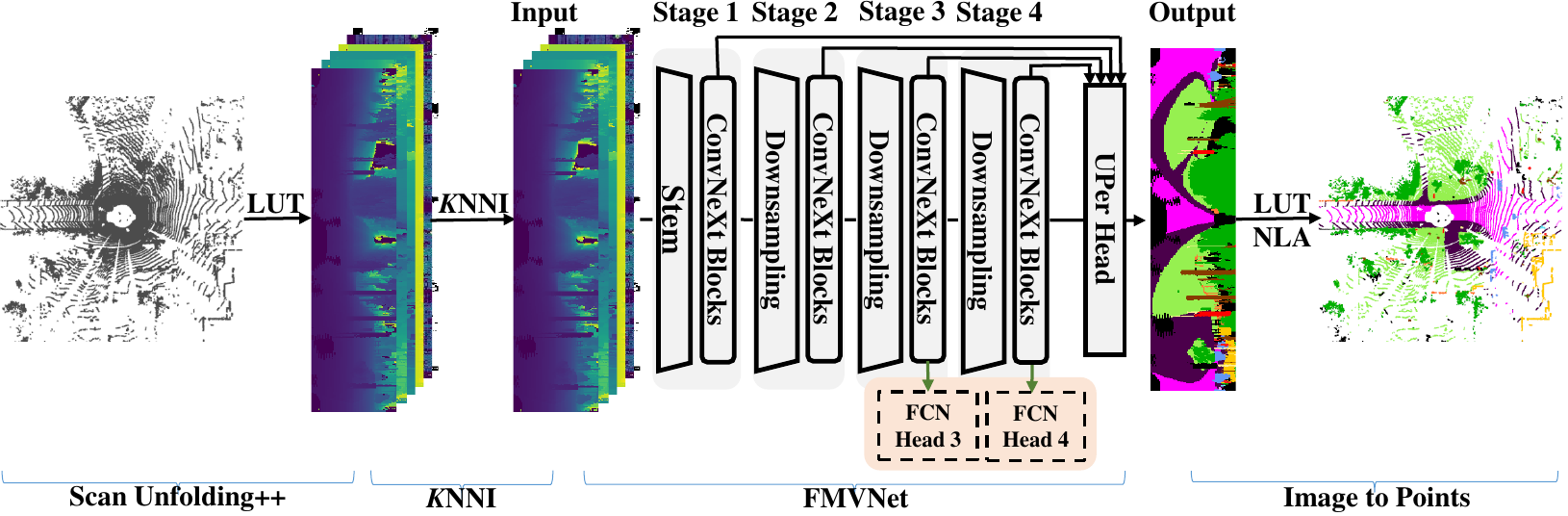}
	\caption{The range images are first generated by the projection method, namely scan unfolding++. Then, we apply range-dependent $K$-nearest neighbor interpolation ($K$NNI) on the images to fill in partial missing points. Subsequently, the images go through the range image-based network, namely FMVNet, to predict the labels. Finally, the outputs are projected back onto the points and pass through the post-processing approach (Nearest Label Assignment~\cite{fidnet_2021}) to obtain the final predictions.}
	\label{fig:overview}
\end{figure*}

To address the above problem, we first propose a novel projection method, namely \textit{scan unfolding++ (SU++)}, to project the points onto the range image. Then we introduce an approach, called \textit{range-dependent $K$-nearest neighbor interpolation ($K$NNI)}, to further fill in the missing values. Finally, we provide a new range image-based model, dubbed \textit{Filling Missing Values Network (FMVNet)}, to achieve state-of-the-art performance in terms of efficiency and accuracy. 

SU++ is different from scan unfolding~\cite{scan_based_projection}, which can only be applied to the raw LiDAR data. In SU++, we provide a new point cloud unfolding algorithm, which is also suitable for deskewing scans~\cite{semantickitti_2019_behley}. Besides, we introduce a ``skewing scans" method to recover the deskewing scans. By SU++, most missing points in the range image are filled in (see the corresponding areas emphasized by the white rectangle and ellipse masks in Fig.~\ref{fig:sp_su_deskew_skew_raw_knni}(c)). Hence, SU++ can effectively avoid the loss of information and increase the upper bounds of segmentation performance.

We propose $K$NNI to fill in the random missing values further. $K$NNI is simple yet effective. By the $K$NNI, all objects look coherent and complete, which can boost the segmentation performance (see Fig.~\ref{fig:sp_su_deskew_skew_raw_knni}(e)). 

FMVNet is also a range image-based model, which builds on ConvNeXt~\cite{convnext2022}. We modify the architecture so as to achieve state-of-the-art performance on LiDAR data. Moreover, we provide a Fast FMVNet by reducing the number of channels and the depth to achieve the better speed-accuracy trade-off.


Extensive experiments conducted on SemanticKITTI~\cite{semantickitti_2019_behley} data show that the proposed SU++ and $K$NNI can significantly improve the performance of existing range image-based models. Also, more experimental results on the SemanticKITTI, SemanticPOSS~\cite{semanticposs_2020}, and nuScenes~\cite{nuscenes_panoptic} datasets validate the effectiveness of the proposed FMVNet and Fast FMVNet.

Our contributions are summarized as follows:
\begin{itemize}
	\item A new projection approach, namely scan unfolding++, is proposed. The range images produced by the scan unfolding++ have fewer missing values. Moreover, the upper bounds of segmentation performance can be raised. 
	
	\item We propose a range-dependent $K$-nearest neighbor interpolation ($K$NNI) method to fill unwanted missing values in the range images. $K$NNI makes objects coherent and complete, thereby boosting the segmentation performance of range image-based models.
	
	\item We introduce the Filling Missing Values Network (FMVNet) and introduce the light version, Fast FMVNet. FMVNet and Fast FMVNet have achieved state-of-the-art performance in terms of efficiency and accuracy. 
\end{itemize}

In the following content, we first discuss related works in Sec.~\ref{sec:related_work}. Then, we provide an overview of the point cloud segmentation pipeline in Sec.~\ref{sec:overview}. Subsequently, we introduce the proposed scan unfolding++ in Sec.~\ref{sec:scan_unfolding++}. We detail the proposed $K$NNI in Sec.~\ref{sec:knni}. We show how to design our FMVNet and Fast FMVNet in Sec.~\ref{sec:fmvnet}. Next, we conduct extensive experiments on the three datasets in Sec.~\ref{sec:experiments}. We provide meaningful discussions in Sec.~\ref{sec:conclusion}. Finally, we conclude our work in Sec.~\ref{sec:conclusion}.

\section{RELATED WORK}\label{sec:related_work}
In this part, we briefly review the previous works related to this paper. 

\subsection{Projection Approaches}
In preparing range images, there are two main projection approaches, namely spherical projection (SP)~\cite{squeezeseg,rangenet++} and scan unfolding (SU)~\cite{scan_based_projection}. SqueezeSeg~\cite{squeezeseg} introduced SP to directly project 3D points onto the 2D range image. Subsequently, almost all range image-based works~\cite{squeezesegv2, squeezesegv3_2020, rangenet++, fidnet_2021, cenet_2022, rangedet_2021, fully_3d_obj_de_2022,epointda2021,rangevit_2023,rangeformer_2023} took this projection method to prepare range images. However, SP causes massive points' occlusion when lasers are not evenly spaced along the vertical direction. To avoid this problem, the work~\cite{scan_based_projection} proposed SU to project each scan line onto each row of the range image. However, the SU algorithm can only be applied to raw LiDAR data. Besides, neither SP nor SU took the negative impact of deskewing scans into consideration. This leads to the missing values along the horizontal direction in the range images. To fill in missing values and avoid the loss of information, we introduce scan unfolding++ (SU++) in this paper. In SU++, we recover the deskewing scans to fill in missing values along the azimuth direction. Moreover, we provide an algorithm to produce ring indices, which are used to unfold the point cloud. SU++ can increase the upper bounds of segmentation performance, thereby improving the performance of existing models. 

\subsection{Interpolation Methods}
Interpolation methods, such as linear, bilinear, nearest neighbor, and moving average, have been widely used in image processing. These approaches are commonly used to resize images. In depth image processing, researchers adopted the interpolation algorithms to correct the estimated depth values and filled in some missing values~\cite{motiondepth2014,correcinter2016,interp2017}. Similarly, this paper introduces an interpolation method, namely range-dependent $K$-nearest neighbor interpolation ($K$NNI), to fill in missing values on the range images. Unlike the commonly used linear and bilinear methods, we directly copy the valid neighbor point with the smallest range to fill in the missing value. This makes more points in the front objects visible and does not introduce noise. More importantly, $K$NNI is simple but can boost segmentation performance.

\subsection{Range Image-based Point Cloud Segmentation} Most point cloud segmentation (PCS) works focus on the design of advanced backbones. For example, SqueezeSeg~\cite{squeezeseg} was the first range image-based approach for the PCS task, where SqueezeNet~\cite{squeezenet_2016} is employed as the backbone. Subsequently, RangeNet++~\cite{rangenet++} adopted the revised DarkNet~\cite{yolov3_2018} as its backbone and introduced a post-processing method, namely k-Nearest-Neighbor search, to refine final predictions. FIDNet~\cite{fidnet_2021} utilized ResNet34~\cite{resnet_2016} as the backbone and designed a fully interpolation decoding module. Afterwards, nearest label assignment (NLA) is proposed to refine the final results further. Based on FIDNet, CENet~\cite{cenet_2022} replaced MLP with convolution, adopted auxiliary branches, and chose more nonlinear activation functions to improve the PCS performance. Recently, RangeViT~\cite{rangevit_2023} and RangeFormer~\cite{rangeformer_2023} took advantage of transformers as their backbones to segment points. Similarly, we introduce the Filling Missing Values Network (FMVNet). It is built on ConvNeXt~\cite{convnext2022}, which has a good speed-accuracy trade-off on ImageNet~\cite{imagenet}. With the advanced backbone, FMVNet can achieve impressive segmentation performance and execution speed. Besides, the light version, Fast FMVNet, can achieve a better speed-accuracy trade-off than existing models. 

\section{Overview}\label{sec:overview}
The pipeline of the point cloud segmentation is provided in Fig.~\ref{fig:overview}. Here, we briefly describe each component in the pipeline. 

The proposed \textit{Scan Unfolding++} (SU++) is first adopted to prepare the range images with fewer missing points. The steps in SU++ are described as follows: (1) LiDAR ring index for each point in a scan is obtained with the proposed Ring Indices Generation (RIG) method. When producing the range images, we use ring indices to unfold the point cloud to avoid missing points along the vertical direction. (2) The ``deskewing" scans are skewed. Using skewing scans aims to avoid dropping points along the horizontal direction in the range image. (3) Look-up table (LUT) is built to efficiently map points to the range image, or vice versa (see Sec.~\ref{sec:scan_unfolding++}).

Then, we propose range-dependent $K$-nearest neighbor interpolation ($K$NNI) to further fill in missing points. The application of $K$NNI is to make the objects in the range images coherent and complete so as to boost the segmentation performance of the models (see Sec.~\ref{sec:knni}).

Subsequently, we introduce a new range image-based segmentation model, dubbed \textit{Filling Missing Values Network} (FMVNet). Missing values in the range images require that a model should have an additional ability to predict them. Hence, considering the speed-accuracy trade-off, we build the FMVNet on the advanced ConvNeXt~\cite{convnext2022}. FMVNet has four stages. The first stage contains Stem and ConvNeXt Blocks. Other stages include Downsampling and ConvNeXt Blocks. Besides, we also use UPer Head~\cite{upernet2018} as the main head and adopt FCN Head~\cite{fcn2015} as auxiliary heads. During the inference phase, all auxiliary heads are dropped. Additionally, we provide a light FMVNet, namely Fast FMVNet, by reducing the model depth and dimension (see Sec.~\ref{sec:fmvnet}).

Finally, we use the LUT to project the predictions back onto the points and apply the nearest neighbor assignment (NLA)~\cite{fidnet_2021} to obtain the final results. Here, NLA can alleviate the problem of point occlusion.

\section{Scan Unfolding++} \label{sec:scan_unfolding++}
In this section, we first elaborate on how to produce the LiDAR ring index for each point in the SemanticKITTI~\cite{semantickitti_2019_behley} dataset. Then, we detail how to skew the ``deskewing" scan. Finally, we show how to make a lookup table to prepare the range image.

\subsection{LiDAR Ring Index}
In this subsection, we equip SemanticKITTI~\cite{semantickitti_2019_behley} data with LiDAR ring indices, which can be utilized to unfold the scans. It is useful for avoiding the missing values along the vertical direction in the generated range images, especially when the lasers in the LiDAR sensor~\cite{hdl_64e_s2_manual} are not vertically spaced. 

\begin{figure}[t]
	\centering
	\includegraphics[width=0.8\columnwidth]{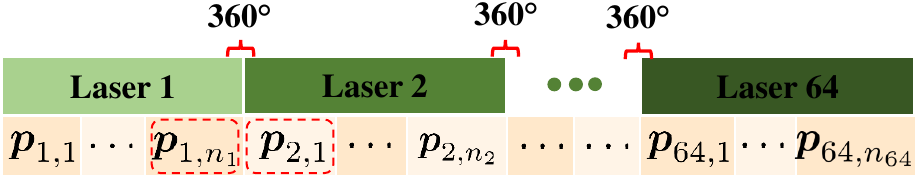}
	\caption{SemanticKITTI~\cite{semantickitti_2019_behley} data representation in a scan. Lasers $1\sim64$ generate $n_1, n_2, \dots, n_{64}$ 3D points, respectively. The azimuth degree gap between the last point (see $\boldsymbol{p}_{1,n_1}$) in the current scan line and the first point in the next scan line (see $\boldsymbol{p}_{2,1}$) is $360^{\circ}$ theoretically.}
	\label{fig:semantic_kitti_data_format}
\end{figure}

The SemanticKITTI~\cite{semantickitti_2019_behley} data format is shown in Fig.~\ref{fig:semantic_kitti_data_format}. There are 64 lasers which produce $n_1, n_2, \dots, n_{64}$ 3D points, respectively. Here $n_1 + n_2 + \dots + n_{64} = N$ and $N$ is the total number of points in a scan. The $n_i, i \in \left\{1, 2, \dots, 64\right\}$ might not be equal. Fortunately, the data has two important properties as follows: (1) Theoretically, the difference between the azimuth degrees of the last point from the laser $i$ and the first point from the laser $i+1$ is $360^{\circ}$ (see the points $\boldsymbol{p}_{1,n_1}$ and $\boldsymbol{p}_{2,1}$ in Fig.~\ref{fig:semantic_kitti_data_format}). (2) All points from the same laser are almost sequentially stored by their azimuth degrees. Only partial points from the same laser do not follow the rule. 

In addition, to produce ring indices for the points, we make the following assumptions: (1) All points in the scan are produced sequentially from the first laser to the last laser; (2) The absolute value between the horizontal angles from two consecutive points in the same scan line is less than a threshold $t$. By the first assumption, we can always assign the points to the ring indices starting from the $0$ even though there might be no points from the first laser. Based on the second assumption, we can guarantee that partial unordered points from the same laser are assigned to the same ring number.

\begin{algorithm}[t]
	\caption{Ring Indices Generation (RIG).}
	\label{alg:ring_indices_generation}
	\small
	\begin{algorithmic}[1]
		\Require All $N$ points $\boldsymbol{P}$ in a scan. A threshold $t$.
		\Ensure All ring indices $\boldsymbol{S}$ corresponding to the points $\boldsymbol{P}$.
		
		\vspace{2ex}
		
		\State $\boldsymbol{x} = \boldsymbol{P}[:, 0]$, $\boldsymbol{y} = \boldsymbol{P}[:, 1]$.
		
		\State $\boldsymbol{\theta} = \text{arctan}(\boldsymbol{y} / \boldsymbol{x}) \times 180^{\circ} / \pi$.
		
		\State $\boldsymbol{m} = \boldsymbol{\theta} < 0$.
		
		\State $\boldsymbol{\theta}[\boldsymbol{m}] = \boldsymbol{\theta}[\boldsymbol{m}] + 360^{\circ}$.
		
		\State Initialize the list $\boldsymbol{S} = [0]$ to store all ring indices.
		
		\State Initialize the first ring index $j = 0$ for the first point.
		
		\For{$i$ in $\{1, 2, \dots N-1\}$}
		
		\State $\theta_i = \boldsymbol{\theta}[i]$, $\theta_{i-1} = \boldsymbol{\theta}[i-1]$.
		
		\If{$\theta_i \geq \theta_{i-1}$ and $|\theta_i - \theta_{i-1}| \leq t$}
		
		\State Append the ring index $j$ to the list $\boldsymbol{S}$.
		
		\Else
		
		\State $j = j + 1$.
		
		\State Append the new ring index $j$ to the list $\boldsymbol{S}$.
		
		\EndIf
		
		\EndFor
	\end{algorithmic}
\end{algorithm}

Based on the data properties and assumptions above, we design a simple yet effective algorithm to produce ring indices for the SemanticKITTI data set. As depicted in Alg.~\ref{alg:ring_indices_generation}, the horizontal angles of all points are calculated (see Lines $1\sim2$), and the range of these angles is from $-180^{\circ}$ to $180^{\circ}$. To make sure that all points from the same laser are sequentially stored and the horizontal angle of the first point theoretically starts from the $0^{\circ}$, we change the negative horizontal angles to the positive values (see Lines $3\sim4$), and now the angles are from $0^{\circ}$ to $360^{\circ}$. In Lines $5\sim6$, we initialize the list to store all ring numbers. Also, we initialize the first ring number. In Line $8$, we get two horizontal angles $\theta_i$ and $\theta_{i-1}$ from two consecutive points. If these two angles satisfy the two conditions, namely ``$\theta_i$ is greater than $\theta_{i-1}$" and ``the difference between them is less than the threshold $t$", we append the current ring number to the list (see Lines $9\sim10$). This means that the current and previous point are from the same laser. Otherwise, the current point is from the next laser (see Lines $12\sim13$). 

In addition, to check whether the Alg.~\ref{alg:ring_indices_generation} generates accurate ring indices for the points in the scan, we propose the following rules: (1) The maximum ring number should be less than $64$ because the SemanticKITTI data is collected by the LiDAR sensor with 64 lasers; (2) According to the experiment, the maximum number of points from the same scan line should be less than or equal to 2180. Based on these two rules, the Alg.~\ref{alg:ring_indices_generation} only makes one inaccurate ring number for the last point in the scan ``002698.bin" in the sequence $13$. The horizontal angle of the last point is $0$, but the $63$ ring indices already exist. Therefore, we see this point as noise and manually label the ring number $63$. After producing all ring indices for all scans in SemanticKITTI, we adopt them to unfold scans.

\subsection{Skewing the ``Deskewing" Scans}\label{sec:skew_the_deskewing_scans}
In this subsection, we recover the ``deskewing" scans to reduce the massive points' occlusion along the horizontal directions when they are projected onto the range image. Here, we build the mathematical model based on the constant velocity model~\cite{loam2014,kissicp2023} because the dataset does not provide other motion estimation data. Besides, for ease of description, we use the name ``deskewing scans" to indicate the scans in SemanticKITTI. We name the recovered scans as ``skewing scans". We use ``raw scans" or ``ground truth scans" to denote the raw LiDAR scans.

In the constant velocity model, the rotational and translational velocities are assumed to be the same as in the previous time step. For the current scan, we denote the angular and translational velocities as $\boldsymbol{\phi}_t$ and $\boldsymbol{v}_t$ at time $t$, respectively. Correspondingly, the rotation matrix and translation vector are expressed as $\boldsymbol{R}_t \in SO(3)$ and $\boldsymbol{t}_t \in \mathbb{R}^3$, respectively. The estimated poses at times $t-1$ and $t-2$ are represented as $\boldsymbol{\zeta}_{t-1} = \left[\begin{array}{cc}
	\boldsymbol{R}_{t-1} & \boldsymbol{t}_{t-1} \\
	\boldsymbol{0}       & 1
\end{array} \right]$ and $\boldsymbol{\zeta}_{t-2} = \left[\begin{array}{cc}
\boldsymbol{R}_{t-2} & \boldsymbol{t}_{t-2} \\
\boldsymbol{0}       & 1
\end{array} \right]$, respectively. Hence, the relative pose $\boldsymbol{\zeta}^{pred}_t$ between the last scan and the current scan can be predicted by the following Eq.~(\ref{eq:relative_pose}):

\begin{align}\label{eq:relative_pose}
	\boldsymbol{\zeta}^{pred}_t & = \boldsymbol{\zeta}^{-1}_{t-2} \boldsymbol{\zeta}_{t-1} \\  \nonumber
	                     & = \begin{bmatrix}
	                     	\boldsymbol{R}_{t-2} & \boldsymbol{t}_{t-2}  \\ \nonumber
	                     	\boldsymbol{0}       & 1 
	                      \end{bmatrix} ^{-1} \begin{bmatrix}
	                      \boldsymbol{R}_{t-1} & \boldsymbol{t}_{t-1}  \\ \nonumber
	                      \boldsymbol{0}       & 1 
                      \end{bmatrix} \\
                      & = \begin{bmatrix}
                      	\boldsymbol{R}_{t-2}^{T} & -\boldsymbol{R}^{T}_{t-2} \nonumber \boldsymbol{t}_{t-2}  \\
                      	\boldsymbol{0}       & 1 
                      \end{bmatrix} \begin{bmatrix}
                      	\boldsymbol{R}_{t-1} & \boldsymbol{t}_{t-1}  \\ \nonumber
                      	\boldsymbol{0}       & 1 
                      \end{bmatrix} \\
                  & = \begin{bmatrix}
                  	\boldsymbol{R}_{t-2}^{T} \boldsymbol{R}_{t-1} & \boldsymbol{R}^{T}_{t-2} (\boldsymbol{t}_{t-1} - \boldsymbol{t}_{t-2}) \\  \nonumber
                  	\boldsymbol{0}       & 1 
                  \end{bmatrix}.
\end{align}
Then, according to Lie theory, we can predict the angular and translational velocities by the following Eqs.~(\ref{eq:angular_velocity}) and~(\ref{eq:translational_velocity}):

\begin{equation}\label{eq:angular_velocity}
	\boldsymbol{\phi}_t = \frac{\text{Log}(\boldsymbol{R}_{t-2}^{T} \boldsymbol{R}_{t-1})}{\triangle t}, 
\end{equation}
\begin{equation}\label{eq:translational_velocity}
	\boldsymbol{v}_t = \frac{\boldsymbol{R}^{T}_{t-2} (\boldsymbol{t}_{t-1} - \boldsymbol{t}_{t-2})}{\triangle t}, 
\end{equation}
where $\text{Log}: SO(3) \rightarrow \mathbb{R}^3$ means the transformation from the manifold space to the corresponding tangent space, and $\triangle t$ indicates the acquisition time of one LiDAR scan. 

During the acquisition time $\triangle t$, we can safely assume that the acquisition time for each point is relative to the acquisition time of the first point. Hence, we can adopt the horizontal angles $\boldsymbol{\theta}$ of the points to calculate the points' relative timestamps $\boldsymbol{\alpha}$. Specifically, for each deskewing point $\boldsymbol{p}_i \in \mathbb{R}^3$, the relative timestamp $\alpha_{i} \in [0, \triangle t]$ can be computed by the following Eq.~(\ref{eq:relative_timestamp}):

\begin{equation}\label{eq:relative_timestamp}
	\alpha_{i} = \frac{\theta_i}{360}, 
\end{equation}
where $\theta_i$ is the horizontal angle of the $i$-th deskewing point $\boldsymbol{p}_i$. Note that here the $\theta_i$ value is in $\left[0, 360\right]$. Finally, the skewing point $\boldsymbol{p}^{\ast}_i$ can be estimated by the Eq.~(\ref{eq:raw_point}) which is:
\begin{equation}\label{eq:raw_point}
	\boldsymbol{p}^{\ast}_i = \text{Exp}(\alpha_{i}\boldsymbol{\phi}_t)^{-1} (\boldsymbol{p}_i - \alpha_{i}\boldsymbol{v}_t),
\end{equation}
where $\text{Exp}: \mathbb{R}^3 \rightarrow SO(3)$ indicates the transformation from the tangent space back to the manifold space.

\subsection{Look-Up Table} 
In this subsection, we build a look-up table (LUT), which is utilized to efficiently project points onto the range image or vice versa. 

Since we have the ring numbers for all points, we can easily generate the corresponding $\boldsymbol{u}$ and $\boldsymbol{v}$ coordinates in the range image by the following Eq.~(\ref{eq:u_v_coords}):
\begin{equation}\label{eq:u_v_coords}
	\left\{
	\begin{split}
		\boldsymbol{u} &= \boldsymbol{\theta} / 360 \times W, \\
		\boldsymbol{v} &= \text{ring numbers}, 
	\end{split} \right.
\end{equation}
where $W$ is the width of the produced range image. Note that here all $\boldsymbol{\theta}$ values are in $\left[0, 360\right]$. Besides, the height of the range image equals to the total number of lasers.

\begin{figure}[t]
	\centering
	\includegraphics[width=0.90\columnwidth]{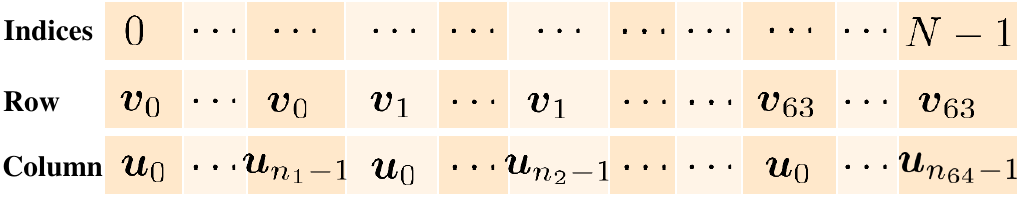}
	\caption{Look-up table for projecting points onto the range image. The first row stores the indices of the points in a scan. The second and third rows store the corresponding $\boldsymbol{v}$ and $\boldsymbol{u}$ coordinates in the range image.}
	\label{fig:look_up_table}
\end{figure}

\begin{figure}[t]
	\centering
	\includegraphics[width=0.75\columnwidth]{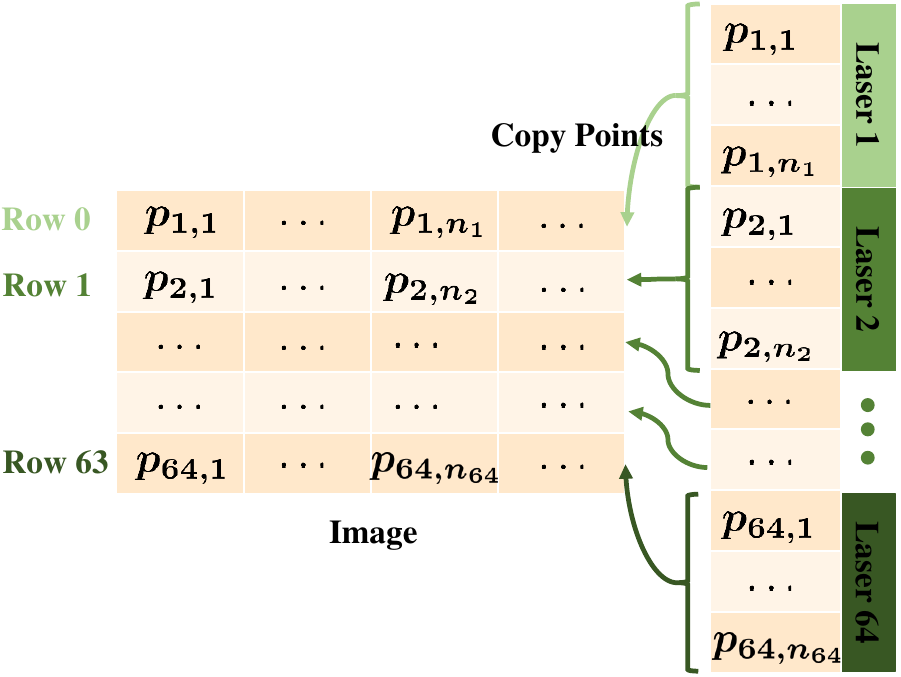}
	\caption{Projection of points onto the range image.}
	\label{fig:project_points_range_image}
\end{figure}

According to the $\boldsymbol{u}$ and $\boldsymbol{v}$ coordinates, we build the LUT (see Fig.~\ref{fig:look_up_table}) where the first, second, and third rows store point indices, as well as the corresponding $\boldsymbol{v}$ and $\boldsymbol{u}$ coordinates, respectively. With the LUT, we can easily project the points onto the range image (see Fig.~\ref{fig:project_points_range_image}). Specifically, a set of points with the same $\boldsymbol{v}$ coordinates are sequentially projected onto the corresponding row of the range image. Inversely, we can return the predicted labels to the point cloud with the LUT.

\subsection{Metrics for Evaluating Scan Unfolding++}\label{sec:metrics_for_scan_unfolding++}
In this subsection, we provide metrics for the evaluation of scan unfolding++.

\subsubsection{Metrics for Skewing Scans}
We first evaluate the proposed constant velocity model used to skew the ``dekewing" scans (see Sec.~\ref{sec:skew_the_deskewing_scans}).  

The purpose of skewing the scan is to recover the raw LiDAR data and avoid the massive points' occlusion along the horizontal direction. Therefore, we propose to measure the difference between the skewing LiDAR data and ground truth LiDAR data. The metrics, namely mean square error (MSE) on the $x$, $y$, $z$ coordinates and the range, are utilized. They are expressed in the Eqs.~(\ref{eq:metric_skew_scan_x_coord}),~(\ref{eq:metric_skew_scan_y_coord}), (\ref{eq:metric_skew_scan_z_coord}), and (\ref{eq:metric_skew_scan_range}): 
\begin{equation}\label{eq:metric_skew_scan_x_coord}
	\text{MSE}_{x} = \frac{1}{N} \sum_{i = 0}^{N-1} (x^{\ast}_i - x^g_i)^2, 
\end{equation}
\begin{equation}\label{eq:metric_skew_scan_y_coord}
	\text{MSE}_{y} = \frac{1}{N} \sum_{i = 0}^{N-1} (y^{\ast}_i - y^g_i)^2, 
\end{equation}
\begin{equation}\label{eq:metric_skew_scan_z_coord}
	\text{MSE}_{z} = \frac{1}{N} \sum_{i = 0}^{N-1} (z^{\ast}_i - z^g_i)^2, 
\end{equation}
\begin{equation}\label{eq:metric_skew_scan_range}
	\text{MSE}_{r} = \frac{1}{N} \sum_{i = 0}^{N-1} (r^{\ast}_i - r^g_i)^2, 
\end{equation}
where $\left\{x^{\ast}_i, y^{\ast}_i, z^{\ast}_i, r^{\ast}_i \right\}$ and $\left\{x^g_i, y^g_i, z^g_i, r^g_i \right\}$ are the coordinates and ranges of the skewing point $\boldsymbol{p}_i^{\ast}$ and ground truth point $\boldsymbol{p}^g_i$. The range is obtained by the $r = \sqrt{x^2 + y^2 + z^2}$.

\subsubsection{Metrics for Kept Points} \label{sec:metrics_for_kept_points}
Using the proposed scan unfolding++ aims to keep as many points as possible in the generated range image. It can also increase the upper bounds of segmentation performance. 

To assess how many projected points are kept in the generated range image, we propose to adopt the ratio of the number of kept points over the total number of points (see Eq.~(\ref{eq:ratio_kept_points})). 
\begin{equation}\label{eq:ratio_kept_points}
	\text{K}_{ratio} = \frac{M}{N}, 
\end{equation}
where $M$ is the number of kept points in the range images, and $N$ is the total number of points.

In addition, we use mean intersection over union (mIoU) to measure the upper bounds of segmentation performance. Here mIoU is expressed in Eq.~(\ref{eq:miou}) which is  
\begin{equation}\label{eq:miou}
	\text{mIoU} = \frac{1}{C}\sum_{i=1}^{C}\text{IoU}_i,
\end{equation}
where ``$\text{IoU}$" is ``$\text{IoU} = \frac{TP}{TP + FP + FN}$"; TP, FP, and FN are true positive, false positive, and false negative predictions, respectively; and $C$ is the total number of classes. Besides, the upper bounds of performance are calculated by the following three steps: First, we project the \textit{points' labels} onto the range images; Second, we project the \textit{pixels' labels} back onto the point cloud; Third, we calculate the mIoU scores (\%) between the \textit{points' labels} and the reprojected \textit{pixels' labels}.

\section{Range-dependent $K$-Nearest Neighbor Interpolation}\label{sec:knni}
In this subsection, we describe the proposed range-dependent $K$-nearest neighbor interpolation ($K$NNI). $K$NNI aims to further fill in the missing values in the range images produced by scan unfolding++ (SU++).

\begin{figure}[t]
	\centering
	\includegraphics[width=0.64\columnwidth]{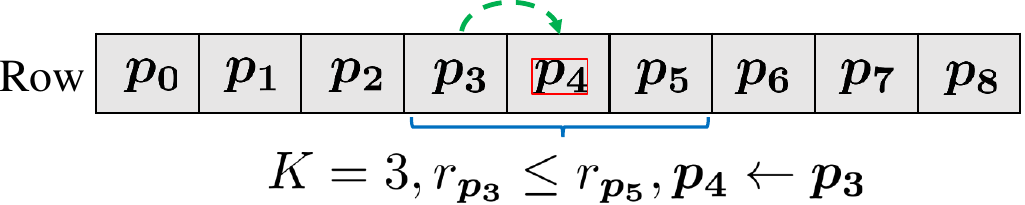}
	\caption{Overview of range-dependent $K$-nearest neighbor interpolation ($K$NNI). Specifically, in a row of the range image, $\boldsymbol{p}_4$ is an ``invalid" pixel's position. If the $K$ is set to 3, $K$NNI will search the nearest 3 positions $\boldsymbol{p}_3$, $\boldsymbol{p}_4$, and $\boldsymbol{p}_5$ in the range image. Then $K$NNI compares the ranges of $\boldsymbol{p}_3$ and $\boldsymbol{p}_5$. If $r_{\boldsymbol{p}_3} \leq r_{\boldsymbol{p}_5}$, the position $\boldsymbol{p}_4$ will be filled with $\boldsymbol{p}_3$.}
	\label{fig:rdknni}
\end{figure}

As illustrated in Fig.~\ref{fig:rdknni}, first, for an ``invalid" pixel's location, its $K$ neighbors within a window in a row of the range image are retrieved. Second, all ``valid" neighbors are compared in terms of their ranges. Finally, the neighbor with the smallest range is selected to fill in the ``invalid" pixel's location. Note that in the first step, we do not consider neighbors from different scan lines because the horizontal angular resolution is typically smaller than the vertical angular resolution. This means that adjacent points in the horizontal direction are commonly closer to each other. Besides, the second step ensures that all points belonging to the front objects are visible~\cite{rangenet++}. The algorithm of $K$NNI is provided in Alg.~\ref{alg:rdknni}.

\begin{algorithm}[t]
	\caption{$K$-Nearest Neighbor Interpolation ($K$NNI).}
	\label{alg:rdknni}
	\small
	\begin{algorithmic}[1]
		\Require A range image $\boldsymbol{I}$ with the size of $\left(H, W, C\right)$. The window size $K$.
		\Ensure A range image $\hat{\boldsymbol{I}}$ processed by $K$NNI.
		
		\vspace{2ex}
		
		\State Get the $\left(\boldsymbol{v}, \boldsymbol{u}\right)$ coordinates of all ``invalid" pixels.
		
		\State Initialize the output range image $\hat{\boldsymbol{I}}$.
		
		\Statex{$\triangleright$ Retrieve neighbor points.}
		
		\For{$s \geq -\frac{K}{2}+1$ and $ s \leq \frac{K}{2}+1$}
		
		\If{s == 0} 
		\State \textbf{Continue}. 
		\EndIf
		
		\State $\boldsymbol{u}_{\text{s}} \gets \boldsymbol{u} + s$. Get horizontal coordinates of neighbors.
		
		\If{some ``invalid" pixels' locations $\left(\boldsymbol{v}_{c}, \boldsymbol{u}_{c}\right)$ are empty}
		
		\Statex{\hspace{\algorithmicindent} \hspace{\algorithmicindent}$\triangleright$ (1) Directly copy the candidate neighbors to fill in the corresponding ``invalid" pixels' locations.}
		
		\State $\hat{\boldsymbol{I}}\left[\boldsymbol{v}_{c}, \boldsymbol{u}_{c}\right] \gets \boldsymbol{I}\left[\boldsymbol{v}_{c}, \boldsymbol{u}_{sc}\right]$. Here the $\left(\boldsymbol{v}_{c}, \boldsymbol{u}_{c}\right)$ are parts of the $\left(\boldsymbol{v}, \boldsymbol{u}\right)$ coordinates and the $\boldsymbol{u}_{sc}$ indicates the horizontal coordinates of the corresponding neighbor points.
		
		\Else
		
		\Statex{\hspace{\algorithmicindent} \hspace{\algorithmicindent} $\triangleright$ (2) Compare the current neighbor points' ranges with the previous ones' ranges and then decide whether the current neighbor points can be used to replace the previous ones in the ``invalid" pixels' locations.}
		
		\State $\boldsymbol{p}_{pre} \gets \hat{\boldsymbol{I}}\left[\boldsymbol{v}_{c}, \boldsymbol{u}_{c}\right]$. Get the previous candidate points.
		
		\State $\boldsymbol{p}_{cur} \gets \boldsymbol{I}\left[\boldsymbol{v}_{c}, \boldsymbol{u}_{sc}\right]$. Get the current candidate points.
		
		\State $\boldsymbol{m} \gets \left( \boldsymbol{r}_{\boldsymbol{p}_{pre}} > \boldsymbol{r}_{\boldsymbol{p}_{cur}} \right)$. Compare the ranges of the previous candidate points with that of the current candidate ones. $\boldsymbol{m}$ is the generated mask. 
		
		\State $\boldsymbol{p}_{pre}[\boldsymbol{m}] \gets \boldsymbol{p}_{cur}[\boldsymbol{m}]$. Replace the previous points.
		
		\State $\hat{\boldsymbol{I}}\left[\boldsymbol{v}_{c}, \boldsymbol{u}_{c}\right] \gets \boldsymbol{p}_{pre}[\boldsymbol{m}]$. Update the range image. 
		
		\EndIf
		
		\EndFor
		
	\end{algorithmic}
\end{algorithm}

In Alg.~\ref{alg:rdknni}, the first step is to find all locations of ``invalid" pixels and their neighbor points. Here ``invalid" pixels are the pixels with zero values. Their locations are expressed by the $\boldsymbol{v}$ and $\boldsymbol{u}$ coordinates (see Line 1). Then, all neighbor points within the window will be traversed. Note that $K$NNI does not process the ``invalid" pixels' locations (see Lines 4$\sim$6). Besides, the neighbor points are searched by adding a step $s$ to the horizontal coordinates $\boldsymbol{u}$ of the ``invalid" pixels (see Line 7). 

The second and third steps are to choose a candidate neighbor point to fill in the ``invalid" pixel location. There are two situations: (1) If the ``invalid" pixel location is not filled with any neighbor point before, the candidate point can be directly copied to the ``invalid" pixel location (see Lines 8$\sim$9). (2) Otherwise, a new candidate point must be compared with the previously copied point in terms of their ranges to decide whether the new one can be used to replace the previous one (see Lines 10$\sim$15).

\section{Filling Missing Values Network}\label{sec:fmvnet}
In this section, we introduce a new range image-based point cloud segmentation model, \textit{i.e.}, Filling Missing Values Network (FMVNet) and its light version, Fast FMVNet. The random missing values require the model to have an additional ability to predict them, so adopting a strong backbone in the segmentation model improves performance. Besides, the model speed is the other important factor to consider in practice. Therefore, we construct FMVNet on the advanced ConvNeXt~\cite{convnext2022} because ConvNeXt has a good speed-accuracy trade-off on ImageNet data~\cite{imagenet} (see Table~\ref{tab:backbones_imagenet}). However, the range image is very different from the colorful image. Hence, in this section, we detail how to revise ConvNeXt towards FMVNet to accommodate the range image. Moreover, we provide a fast version, Fast FMVNet, to achieve a higher execution speed.

\begin{table*}[t]
	\caption{Modifications of ConvNeXt~\cite{convnext2022} towards FMVNet. The size of the input image is set to $6 \times 64 \times 2048$. KS: kernel size. S: stride. Param.: the number of model parameters. FPS: frames per second.}
	\label{tab:convnext_to_fmvnet}
	\centering
	\scalebox{1.0}{
		\begin{tabular}{l|c|c|c|c}
			\hline
			Modifications      & Param.   & FLOPs     & FPS    & mIoU            \\ \hline \hline
			No changes         & 59.26M   & 1278.78G  & 24.64  & 59.5            \\ \cline{2-5}    
			In Stem, KS: $4\times4$, S: $4\times4$ $\rightarrow$ KS: $1\times1$, S: $1\times1$  & 59.25M & 1869.53G & 10.41  & 68.0 \\ \cline{2-5}
			Auxiliary Heads 3 and 4; Weight 0.4   & 59.25M & 1869.53G & 10.41  & 68.6 \\ \hline 
	\end{tabular}}  
\end{table*}

\begin{figure*}[t]
	\centering
	\includegraphics[width=1.63\columnwidth]{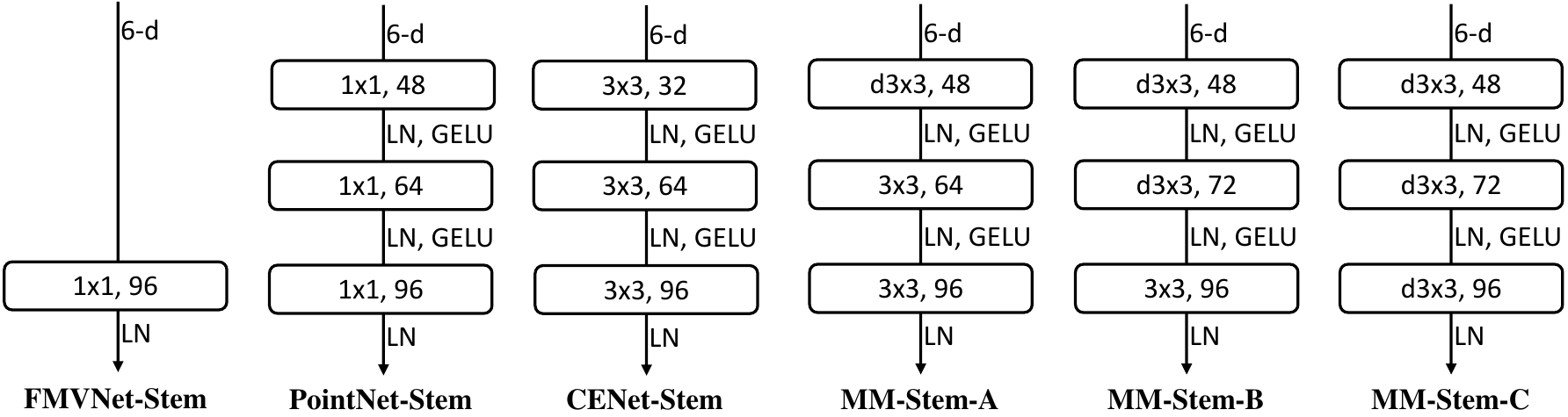}
	\caption{The designs of FMVNet-Stem, PointNet-Stem~\cite{pointnet_2017}, CENet-Stem~\cite{cenet_2022}, and MM-Stem-A/B/C. ``d$3\times3$": the depthwise convolution with the kernel size of $3\times3$. MM: multi-modal.}
	\label{fig:stem_modules}
\end{figure*}

\begin{table}[t]
	\caption{Comparisons among various image classification models on ImageNet-1K~\cite{imagenet}. The image size is set to $224\times224$. All models are tested on one GeForce RTX 3080 GPU. Param.: the number of model parameters. FPS: frames per second. Top-1: top-1 accuracy (\%).}
	\label{tab:backbones_imagenet}
	\centering
	\scalebox{1.0}{
		\begin{tabular}{l|c|c|c|c|c}
			\hline
			Models                              & Years  & Param.    & FLOPs  & FPS   & Top-1    \\ \hline   \hline  
			Swin-T~\cite{swintrans2021}         & 2021  & 28.3M     & 4.5G   & 128.9 & 81.2     \\ \hline
			PVTv2-B2~\cite{pvtv22022}           & 2022  & 25.4M     & 4.0G   & 92.0  & 82.0     \\ \hline
			CSWin-T~\cite{cswin2022}            & 2022  & 22.3M     & 4.3G   & 33.3  & 82.8     \\ \hline
			ConvNeXt-T~\cite{convnext2022}      & 2022  & 28.6M     & 4.5G   & 224.9 & 82.1     \\ \hline
			F-Swin-T~\cite{flattentrans2023}    & 2023  & 29.2M     & 4.5G   & 104.0 & 82.1     \\ \hline
			F-PVTv2-B2~\cite{flattentrans2023}  & 2023  & 22.6M     & 4.3G   & 56.8  & 82.5     \\ \hline
			F-CSwin-T~\cite{flattentrans2023}   & 2023  & 21.2M     & 4.3G   & 33.6  & 83.1     \\ \hline
			TransNeXt-T~\cite{transnext2023}    & 2024  & 28.3M     & 5.7G   & 32.8  & 84.0     \\ \hline 
	\end{tabular}}  
\end{table}

\subsection{ConvNeXt-Tiny} 
In this subsection, we detail the architecture specifications of ConvNeXt-Tiny~\cite{convnext2022}.

It includes four stages. The first stage contains a stem module and ConvNeXt Blocks. Other stages consist of a downsampling module and ConvNeXt Blocks. In the stem module, authors used a convolution with the kernel size of $4\times4$ and the stride of $4\times4$, as well as a layer normalization to significantly decrease the input image size to reduce computational cost and the redundant information. The ConvNeXt Block is constituted of a depthwise convolution with the $7\times7$ kernel size, a layer normalization, an activation layer, and two linear layers. The downsampling module comprises a layer normalization and a convolution with the kernel size of $2\times2$ and the stride of $2\times2$ to reduce the size of feature maps. Besides, the depths and channels for the four stages are set to $\left[3, 3, 9, 3\right]$ and $\left[96, 192, 384, 768\right]$, respectively. 

In addition, on the semantic image segmentation task, authors adopted UPer Head~\cite{upernet2018} as the main head and utilized FCN Head~\cite{fcn2015} as the auxiliary head. The auxiliary head is attached to the stage 3. The weight for the corresponding auxiliary loss is set to 0.4.

\subsection{FMVNet} 
Here, we slightly modify ConvNeXt-Tiny to become FMVNet. 

\subsubsection{Stem Module} 
We change the kernel size of $4\times4$ and the stride of $4\times4$ in the stem module to $1\times1$ and $1\times1$ (see FMVNet-Stem in Fig.~\ref{fig:stem_modules}). The reasons are as follows: (1) Different from the natural image, the range image lacks colors and might not have redundancy. (2) The height of the range image is only 64 on SemanticKITTI data~\cite{semantickitti_2019_behley}. Reducing the image size dramatically by four times severely decreases the segmentation performance. By FMVNet-Stem, the mIoU score (\%) is increased considerably from 59.5 to 68.0 (see Table.~\ref{tab:convnext_to_fmvnet}). To further validate the effectiveness of FMVNet-Stem, we discuss other choices, namely PointNet-Stem~\cite{pointnet_2017}, CENet-Stem~\cite{cenet_2022}, MM-Stem-A/B/C, in the experiments (see Sec.~\ref{sec:stem_variants}). 

\subsubsection{Auxiliary Heads \& Weights}
The auxiliary heads aim to provide extra supervision for FMVNet during the training phase to boost its performance. For ConvNeXt-Tiny~\cite{convnext2022} on the semantic image segmentation task, only one auxiliary head is attached to the stage 3, and the corresponding weight is set to 0.4. Here, we add an extra auxiliary head to the stage 4 during the training phase. In experiments (see Sec.~\ref{sec:auxiliary_heads_weights}), we find that this setting increases the mIoU score (\%) from 68.0 to 68.6.

\subsubsection{Normalization}\label{sec:normalization}
We also discuss the removal of the normalization layer after each stage. However, removing these layers decreases the segmentation performance (see Sec.~\ref{sec:remove_normalization}). Until now, the model can achieve high segmentation performance on the SemanticKITTI validation dataset. And we call it FMVNet. 

\subsection{Fast FMVNet}\label{sec:fast_fmvnet}
In this subsection, we describe the Fast FMVNet. 

\begin{table}[t]
	\caption{The exploration of depths and channels of Fast FMVNet in terms of model parameters (Param.), FLOPs, frames per second (FPS), and mIoU scores (\%).}
	\label{tab:fmvnet_depth_channels}
	\centering
	\scalebox{0.88}{
		\begin{tabular}{c|c|c|c|c|c}
			\hline
			Depths       & Channels            & Param.   & FLOPs    & FPS   & mIoU    \\ \hline \hline  
			[3, 3, 9, 3] & [96, 192, 384, 768] & 59.25M   & 1869.87G & 15.44 & 68.3    \\ \hline
			[3, 3, 9, 3] & [128, 128, 128, 128]& 4.58M    & 189.47G  & 48.67 & 66.9    \\ \hline
			[3, 4, 6, 3] & [128, 128, 128, 128]& 4.31M    & 190.60G  & 48.10 & 67.4    \\ \hline
	\end{tabular}}  
\end{table}

We first make the tensor shape consistent in FMVNet. The tensor shape can be expressed by $\left[B, H, W, C\right]$ where $B$, $H$, $W$, and $C$ mean the batch size, height, width, and the number of channels. The transformation between the tensor shape $\left[B, H, W, C\right]$ and $\left[B, C, H, W\right]$ decreases the speed. To keep the same tensor shape, we change all layer normalization to batch normalization and slightly modify the architecture. This helps FMVNet to increase the speed from 10.41 FPS to 15.44 FPS (see Table~\ref{tab:fmvnet_depth_channels}) while obtaining the 68.3\% mIoU score. 

We further increase the speed by reducing the number of channels in FMVNet. Similar to the settings in FIDNet~\cite{fidnet_2021} and CENet~\cite{cenet_2022}, we decrease the numbers of channels in the four stages to $\left[128, 128, 128, 128\right]$. Correspondingly, we change the dimension in UPer Head to 128. By these modifications, the speed of FMVNet is raised to 48.67 FPS, and the model can achieve the 66.9\% mIoU score (see Table~\ref{tab:fmvnet_depth_channels}). Moreover, same as the settings in FIDNet and CENet, we further reduce the numbers of ConvNeXt blocks in the four stages from $\left[3, 3, 9, 3\right]$ to $\left[3, 4, 6, 3\right]$. The speed of FMVNet is slightly decreased, \textit{i.e.}, from 48.67 FPS to 48.10 FPS, but the model can obtain the improved performance (\textit{i.e.}, 67.4\% mIoU score). Considering the balance between the speed and mIoU score, we define the Fast FMVNet with the depths of $\left[3, 4, 6, 3\right]$ and the channels of $\left[128, 128, 128, 128\right]$. Note that compared with FMVNet, Fast FMVNet only has 4.31M parameters but achieves the 67.4\% mIoU score. The experimental results will validate the effective design of Fast FMVNet. 

\subsection{Loss Function}
Following previous works~\cite{cenet_2022,litehdseg2021}, we adopt the combination of weighted cross-entropy loss $L_{\text{wce}}$, lov\'{a}sz-softmax loss $L_{\text{ls}}$, and boundary loss $L_{\text{bd}}$. The total loss function contains the losses from the main head $L^{\text{main}}$ and auxiliary heads $L^{\text{auxiliary}}$, which is expressed in Eq.~(\ref{eq:loss_funcs}),
\begin{equation}\label{eq:loss_funcs}
	\text{Loss} = L^{\text{main}} + w_4 \times \sum_{i=0}^{1} L^{\text{auxiliary}}_i,
\end{equation}
where $L = w_1 \times L_{\text{wce}} + w_2 \times L_{\text{ls}} + w_3 \times L_{\text{bd}}$ and the weights $w_1$, $w_2$, $w_3$, and $w_4$ are set to 1, 1, 1.5, and 0.4.

\section{EXPERIMENTS}\label{sec:experiments}
In this section, we first explain experimental settings. Then, we draw comparisons among range image generation methods. Subsequently, we compare popular point cloud segmentation models with their counterparts trained on the range images produced by the proposed scan unfolding++ and range-dependent $K$-nearest neighbor interpolation ($K$NNI). In the next, we provide the ablation study of the proposed Filling Missing Values Network (FMVNet). Finally, we show more experimental results on the SemanticKITTI~\cite{semantickitti_2019_behley}, SemanticPOSS~\cite{semanticposs_2020}, and nuScenes~\cite{nuscenes_panoptic} datasets.

\subsection{Experimental Settings}
\subsubsection{Datasets}
We conducted experiments on SemanticKITTI~\cite{semantickitti_2019_behley}, SemanticPOSS~\cite{semanticposs_2020}, and nuScenes~\cite{nuscenes_panoptic} data sets. \textbf{SemanticKITTI} is a large-scale and high-quality point cloud dataset which provides per-point labels. In it, the sequences $\left\{00 \sim 07, 09 \sim 10\right\}$, $\left\{08\right\}$, and $\left\{11 \sim 21\right\}$ are served as the training, validation, and test data sets, respectively. Besides, only 19 classes are considered under the condition of a single scan. In addition, the dataset provides poses and timestamps corresponding to LiDAR scans. Moreover, the raw LiDAR data, sequences $\left\{00, 01, 02, 04, 05, 06, 07, 08, 09, 10\right\}$, can be used to validate the effectiveness of the proposed skewing scan method. \textbf{SemanticPOSS} contains six sequences $\left\{00 \sim 05\right\}$ in which the sequence $\left\{02\right\}$ serves as the test dataset and the rest is the training data. Furthermore, 14 classes are labelled. Besides, the dataset provides tags with which we can easily get the ring numbers. \textbf{nuScenes} is also a large-scale outdoor point cloud dataset. It includes 28,130 training, 6,019 validation, and 6,008 test scans. Besides, only 16 semantic classes are considered. Moreover, the data set contains ring numbers in each scan. 

\subsubsection{Models and Implementation Details} We adopted three popular range image-based point cloud segmentation models, \textit{i.e.}, RangeNet53++~\cite{rangenet++}, FIDNet~\cite{fidnet_2021}, and CENet~\cite{cenet_2022} in our experiments for fair comparison because these models are open-source and reproducible. 

Besides, we utilized the data augmentation techniques~\cite{rangenet++,fidnet_2021,cenet_2022,rangeformer_2023} such as random scaling, random horizontal flip, random rotation, PolarMix~\cite{polarmix_2022}, and LaserMix~\cite{lasermix2023} to train the models. The batch size was set to 16 for RangeNet53++, FIDNet, and CENet on SemanticKITTI~\cite{semantickitti_2019_behley}, SemanticPOSS~\cite{semanticposs_2020}, and nuScenes~\cite{nuscenes_panoptic} datasets. The batch size was set to 8 for our FMVNet and Fast FMVNet. We trained all models on a server with 4 NVIDIA A100 GPUs. In addition, to respect ConvNeXt~\cite{convnext2022}, we fixed all random seeds to ``123" during the training and testing phases for reproduction and fair comparisons. The learning rate and weight decay were set to 0.002 and 0.0001, respectively. We adopted the AdamW optimizer to train the models. Moreover, the intersection-over-union (IoU) score over each class and the mean IoU (mIoU) score over all classes were reported.

\subsection{Comparisons among Range Image Generation Methods} 
In this subsection, we first validated the effectiveness of the proposed skewing scan method. Then, we compared the proposed scan unfolding++ with the commonly used spherical projection in terms of how many points are kept, the upper bounds of performance, and the performance gains of the range image-based models. 

\subsubsection{Skewing Scans} 
The aim of skewing the scans is to avoid missing points along the horizontal direction when projected onto the range image. Note that we conducted experiments only on the sequences $\left\{00, 01, 02, 04, 05, 06, 07, 08, 09, 10\right\}$ because only these sequences include raw LiDAR data. According to the metrics in the section~\ref{sec:metrics_for_scan_unfolding++}, we provided the results of $\text{MSE}_{x}$, $\text{MSE}_{y}$, $\text{MSE}_{z}$, and $\text{MSE}_{r}$ in Table~\ref{tab:skewing_the_scan}.

\begin{table}[t]
	\caption{Comparisons between the skewing scans and deskewing scans in terms of $\text{MSE}_{x}$, $\text{MSE}_{y}$, $\text{MSE}_{z}$, and $\text{MSE}_{r}$ values. ``Seq.": sequence; ``SK": skewing; ``DSK": deskewing.}
	\label{tab:skewing_the_scan}
	\centering
	\scalebox{0.84}{
		\begin{tabular}{c|r|r|r|r|r|r|r|r}
	    \hline   
	    \multirow{2}*{Seq.} & \multicolumn{2}{c|}{$\text{MSE}_{x} (\times 10^{-4})$} & \multicolumn{2}{c|}{$\text{MSE}_{y} (\times 10^{-4})$} &  \multicolumn{2}{c|}{$\text{MSE}_{z} (\times 10^{-4})$} & \multicolumn{2}{c}{$\text{MSE}_{r} (\times 10^{-4})$} \\ \cline{2-9}
	              & SK   & DSK    & SK    & DSK   & SK  & DSK   & SK    & DSK      \\ \hline \hline
	    00        & 4.1  & 627.7  & 3.6   & 55.6  & 0.3 & 2.8   & 3.1   & 314.1   \\ \hline
	    01        & 43.5 & 4053.3 & 8.0   & 54.7  & 0.2 & 1.5   & 30.7  & 2110.2  \\ \hline
	    02        & 9.6  & 1005.1 & 3.9   & 48.4  & 0.1 & 2.4   & 8.0   & 520.7   \\ \hline
	    04        & 5.9  & 1685.0 & 0.2   & 0.1   & 0.1 & 1.2   & 3.3   & 871.1   \\ \hline
	    05        & 3.6  & 593.9  & 2.3   & 30.9  & 0.1 & 1.6   & 2.6   & 303.9   \\ \hline
	    06        & 5.1  & 1097.8 & 0.9   & 71.3  & 0.1 & 1.5   & 2.8   & 547.3   \\ \hline
	    07        & 2.6  & 431.1  & 1.0   & 48.3  & 0.1 & 1.4   & 1.9   & 211.3   \\ \hline
	    08        & 5.3  & 592.6  & 4.6   & 50.0  & 0.9 & 2.9   & 3.9   & 296.6   \\ \hline
	    09        & 12.4 & 996.2  & 6.1   & 55.4  & 0.2 & 1.9   & 10.4  & 512.6   \\ \hline
	    10        & 5.9  & 576.7  & 4.9   & 35.8  & 0.2 & 2.2   & 4.9   & 295.5   \\ \hline
	\end{tabular}}
\end{table}

Table~\ref{tab:skewing_the_scan} shows that the recovered LiDAR data (skewing scans) is almost the same as the raw LiDAR data in terms of MSE values (see all SK columns). Only on the sequence $\left\{01\right\}$, there is a relatively large discrepancy between the recovered LiDAR data and the ground truth LiDAR data due to some non-constant velocity. Specifically, the MSE values on the $x$ coordinates and the ranges are 0.00435 and 0.00307, respectively. However, these values are still too small compared with the counterparts computed between the ground truth LiDAR data and deskewing data (\textit{i.e.}, 0.00435 vs. 0.40533 and 0.00307 vs. 0.21102). Besides, the image (c) in Fig.~\ref{fig:sp_su_deskew_skew_raw_knni} from the skewing scan and the image (d) from the raw scan are almost the same. Therefore, the experimental results validate the effectiveness of the proposed skewing scan method.

\subsubsection{Scan Unfolding++ vs. Spherical Projection} 
Compared with the commonly used spherical projection (SP), the purpose of the proposed scan unfolding++ (SU++) is to keep more points in the generated range image so as to reduce the loss of information. Hence, we first estimated how many projected points are stored in the range images with the proposed $\text{K}_{ratio}$ metric. Then, we computed the upper bounds of segmentation performance (see Sec.~\ref{sec:metrics_for_kept_points}). The experimental results were provided in Table~\ref{tab:su++_sp}.

\begin{table}[t]
	\caption{Comparisons between the proposed scan unfolding++ (SU++) and spherical projection (SP) in terms of the $\text{K}_{ratio}$ and mIoU scores (\%). ``+ DSK" means the projection method applied to the deskewing scans. ``+ SK" indicates the projection approach employed to the skewing scans. ``+ Raw" shows the projection method used for the raw LiDAR scans.}
	\label{tab:su++_sp}
	\centering
	\scalebox{0.94}{
		\begin{tabular}{l|c|c|c|c|c|c}
			\hline   
			\multirow{2}*{Methods}  & \multicolumn{2}{c|}{$64\times512$} & \multicolumn{2}{c|}{$64\times1024$} & \multicolumn{2}{c}{$64\times2048$}  \\ \cline{2-7}
			& $\text{K}_{ratio}$ & mIoU      & $\text{K}_{ratio}$  & mIoU    & $\text{K}_{ratio}$   & mIoU       \\ \hline \hline
			SP+DSK   & 20.98 & 79.09 & 41.01 & 84.77  & 77.46  & 88.81   \\ \hline
			SP+SK    & 21.05 & 80.31 & 41.29 & 86.69  & 79.22  & 91.20   \\ \hline
			SP+Raw   & 21.00 & 80.31 & 41.22 & 86.71  & 79.09  & 91.24   \\ \hline \hline
			SU+++DSK & 24.11 & 82.64 & 47.47 & 89.12  & 89.47  & 93.49   \\ \hline
			SU+++SK  & 24.18 & 84.69 & 47.92 & 92.58  & 92.15  & 97.96   \\ \hline
			SU+++Raw & 24.14 & 84.74 & 47.89 & 92.67  & 92.16  & 98.05   \\ \hline   
	\end{tabular}}
\end{table}

Note that in Table~\ref{tab:su++_sp}, for the methods ``SP+DSK", ``SP+SK", ``SU+++DSK", and ``SU+++SK", we computed the $\text{K}_{ratio}$ scores (\%) on the whole SemanticKITTI dataset. Also, we calculated the upper bounds of performance (mIoU scores (\%)) only on the training and validation datasets, as there are no labels in the test data set. Besides, for the approaches ``SP+Raw" and ``SU+++Raw", we computed the $\text{K}_{ratio}$ and mIoU scores (\%) only on the sequences $\left\{00, 01, 02, 04, 05, 06, 07, 08, 09, 10\right\}$. Besides, we set the sizes of the range image to $64\times512$, $64\times1024$, and $64\times2048$.

Table~\ref{tab:su++_sp} shows that SU++ can keep more points in the produced range images compared with SP under various sizes. Correspondingly, the upper bounds of performance by SU++ are higher than the counterparts by SP. This is because SU++ can avoid the massive points' occlusion in the range image (see the images (a), (b), and (c) in Fig.~\ref{fig:sp_su_deskew_skew_raw_knni}). It also means that SU++ can help reduce the loss of information. In addition, comparing ``SP+DSK" with ``SP+SK" and comparing ``SU+++DSK" with ``SU+++SK", we found that skewing scans help avoid the points' overlap in the horizontal direction in the range image (see the image (c) in Fig.~\ref{fig:sp_su_deskew_skew_raw_knni}). Correspondingly, the skewing scans have higher upper bounds of performance than the deskewing scans. Moreover, we saw that under different range image sizes, ``SP+SK" achieves almost the same $\text{K}_{ratio}$ and mIoU scores (\%) as ``SP+Raw". ``SU+++SK" and  ``SU+++Raw" also obtains the similar $\text{K}_{ratio}$ and mIoU scores (\%). These results further validate the effectiveness of the proposed skewing scan method. More importantly, under the image size of $64\times2048$, the 97.96\% mIoU score of ``SU+++SK" surpasses the 88.81\% mIoU score of ``SP+DSK" by a large margin. We will see that the significantly increased upper bound of performance leads to the performance gains of existing range image-based segmentation models. The experimental results prove the effectiveness of the proposed SU++.

\begin{figure}[t]
	\centering
	\includegraphics[width=0.97\columnwidth]{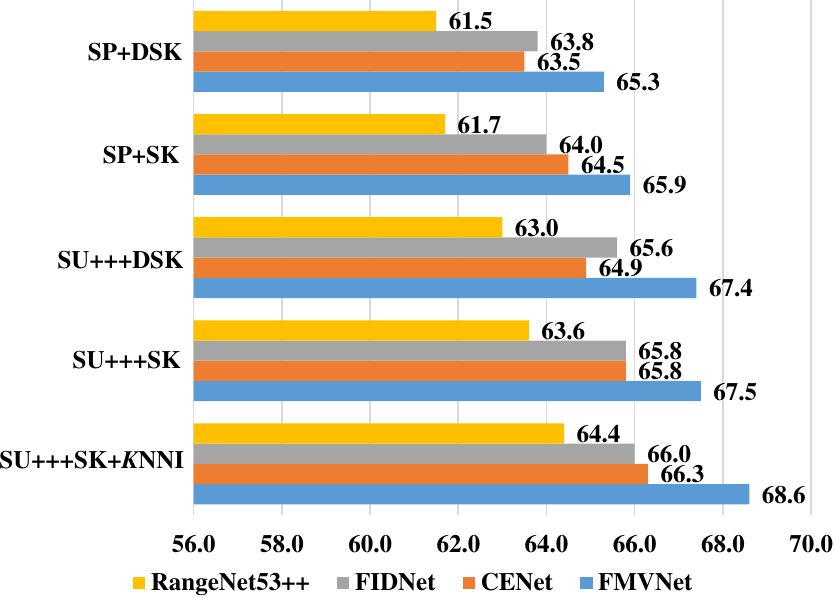}
	\caption{RangeNet53++, FIDNet, CENet, and our FMVNet were trained on ``SP+DSK", ``SP+SK", ``SU+++DSK", ``SU+++SK", ``SU+++SK+$K$NNI" based images. The results were reported on the SemanticKITTI validation dataset. Note that all models were trained with the fixed random seed, and \textbf{NO TTA} is applied to the results.}
	\label{fig:sp_su_dsk_sk_knni}
\end{figure}

\subsubsection{Models Trained on SU++ and SP Based Images}\label{sec:model_on_su++_sp}
To further compare the proposed scan unfolding++ (SU++) and spherical projection (SP), we trained RangeNet53++~\cite{rangenet++}, FIDNet~\cite{fidnet_2021}, CENet~\cite{cenet_2022}, and our FMVNet on SU++ and SP based images, respectively. Then, we reported the mIoU scores (\%) on the SemanticKITTI~\cite{semantickitti_2019_behley} validation dataset. The experimental results were described in Fig.~\ref{fig:sp_su_dsk_sk_knni} and Table~\ref{tab:64x2048_kitti_val_results}. Here, ``SP+DSK" means the spherical projection on deskewing scans. ``SP+SK" indicates the spherical projection on skewing scans. ``SU+++DSK" denotes the scan unfolding++ on deskewing scans. ``SU+++SK" means the scan unfolding++ on skewing scans. 

In Fig.~\ref{fig:sp_su_dsk_sk_knni}, we saw that by SU++, the mIoU score of RangeNet53++ is increased from 61.5\% to 63.6\%. The mIoU score of FIDNet is increased to 65.8\%. CENet obtains a higher mIoU score than the baseline (65.8\% vs. 63.5\%). The mIoU score of our FMVNet is also raised to 67.5\% after training on the ``SU+++SK" based images. This proves that fewer missing values lead to higher bounds of performance and further boost the performance of segmentation models.

\subsection{Models Trained with $K$NNI}\label{sec:model_on_knni}
The range-dependent $K$-nearest neighbor interpolation ($K$NNI) is proposed to further fill in missing values in the range images to make objects coherent and complete (see Sec.~\ref{sec:knni}). In this subsection, we validated the effectiveness of $K$NNI. In the experiments, we trained RangeNet53++~\cite{rangenet++}, FIDNet~\cite{fidnet_2021}, CENet~\cite{cenet_2022}, and our FMVNet on ``SU+++SK+$K$NNI" based images, and reported results on the SemanticKITTI~\cite{semantickitti_2019_behley} validation dataset. Here ``SU+++SK+$K$NNI" means that $K$NNI applied to the range images generated by scan unfolding++ on the skewing scans. The experimental results were shown in Fig.~\ref{fig:sp_su_dsk_sk_knni} and Table~\ref{tab:64x2048_kitti_val_results}. We saw that with the proposed $K$NNI, the mIoU scores of RangeNet53++, FIDNet, CENet, and FMVNet are further increased to 64.4\%, 66.0\%, 66.3\%, and 68.6\%, respectively. The performance gains validate the effectiveness of the proposed $K$NNI.

In addition, we provided two other options for comparison. 

\subsubsection{$K$NNI-A} In $K$NNI, we directly copied the neighbor point with the smallest range to fill in the ``invalid" pixel position. This is consistent with the work~\cite{rangenet++} where authors sorted all points based on their ranges to make the front objects visible in the range image. For ease of description, we here named $K$NNI as $K$NNI-A.

\begin{table}[t]
	\caption{Comparisons among $K$NNI-A, $K$NNI-B, and $K$NNI-C in terms of mIoU scores (\%).}
	\label{tab:knni_ABC}
	\centering
	\scalebox{1.0}{
		\begin{tabular}{l|c|c|c}
			\hline   
			Models       & $K$NNI-A & $K$NNI-B & $K$NNI-C  \\ \hline \hline
			RangeNet53++ & 64.41    & 64.38    & 62.33     \\ \hline
			FIDNet       & 65.97    & 65.80    & 63.14     \\ \hline
			CENet        & 66.32    & 65.24    & 63.73     \\ \hline
			FMVNet       & 68.62    & 67.65    & 66.65     \\ \hline
	\end{tabular}}
\end{table}

\subsubsection{$K$NNI-B}
In $K$NNI-B, we used the mean value over all neighbor points to fill in the ``invalid" pixel position. However, we still adopted the label of the neighbor point with the smallest range to serve as the label of the ``invalid" pixel. Using the mean value can smooth the input data, but this can make some noise because the new point might not fall on any objects.  

\subsubsection{$K$NNI-C}
In $K$NNI-C, we still used the point with the smallest range to fill in the ``invalid" pixel position. However, if the labels from the left and right neighbors are different, we set the label of the new point to the ``ignored" label. During the training phase, we did not compute the loss on the ``ignored" labels. This can avoid the confusion as to what label should be assigned to the boundary ``invalid" pixel.

\subsubsection{Analysis of $K$NNI-A/B/C}
The comparison results were provided in Table~\ref{tab:knni_ABC}. We saw that all models with $K$NNI-A can achieve the highest mIoU scores (\%) compared with their counterparts. By comparing $K$NNI-A and $K$NNI-B, we found that the noise data caused by the average value over neighbors slightly degenerates the performance. By comparing $K$NNI-A and $K$NNI-C, we can safely conclude that explicitly processing the boundary ``invalid" pixels cannot improve the performance. We guessed that the models recognize objects by their boundaries in LiDAR data. For boundary pixels in the range image, valid inputs with "ignored" labels might confuse the models during the training phase, thereby leading to inferior performance. Based on the comparison results, we validated the effectiveness of the proposed $K$NNI-A.

\subsection{Ablation Study for FMVNet} 
In this subsection, we first discussed other choices about the number of auxiliary heads and corresponding weights. Then, we made a comparison among various stem modules. Finally, we explored the removal of the layer normalization after four stages.

\subsubsection{Auxiliary Heads \& Weights}\label{sec:auxiliary_heads_weights}
The auxiliary heads aim to provide extra supervision for FMVNet during the training phase to boost model performance. For ConvNeXt~\cite{convnext2022} on the semantic image segmentation task, only one auxiliary head is attached to the stage 3, and the corresponding weight is set to 0.4. By contrast, in CENet~\cite{cenet_2022}, authors appended more auxiliary heads to the stages 2, 3, and 4, and set the weight to 1.0. In this section, we compared the models with different settings of auxiliary heads and weights. The experimental results on the SemanticKITTI~\cite{semantickitti_2019_behley} validation set were reported in Table~\ref{tab:fmvnet_aux}. We saw that adding auxiliary heads to the stages [3, 4] and setting the weight to 0.4 improve the model performance.

\begin{table}[t]
	\caption{Different settings of auxiliary heads and weights for FMVNet during the training phase.}
	\label{tab:fmvnet_aux}
	\centering
	\scalebox{1}{
		\begin{tabular}{l|c|c}
			\hline
			Auxiliary Heads  & Weights & mIoU     \\ \hline \hline  
			To Stage  [3]              & 0.4        & 68.0     \\ \hline
			To Stages [3, 4]           & 0.4        & 68.6     \\ \hline
			To Stages [2, 3, 4]        & 0.4        & 67.1     \\ \hline
			To Stages [1, 2, 3, 4]     & 0.4        & 67.6     \\ \hline
			
			To Stage  [3]              & 1.0        & 67.9     \\ \hline
			To Stages [3, 4]           & 1.0        & 67.8     \\ \hline
			To Stages [2, 3, 4]        & 1.0        & 67.9     \\ \hline
			To Stages [1, 2, 3, 4]     & 1.0        & 67.7     \\ \hline
	\end{tabular}}  
\end{table}

\subsubsection{Stem Modules}\label{sec:stem_variants}
In this subsection, we compared different stem modules. The stem module aims to transform the inputs into the feature maps. In our FMVNet, the stem consists of a convolution with the kernel size of $1\times1$ and a layer normalization. The output feature maps have the same size as inputs and have 96 channels (see Sec.~\ref{sec:fmvnet}). For the ease of description, we named our stem module as FMVNet-Stem (see Fig.~\ref{fig:stem_modules}). In the following content, we described the alternatives. 

\begin{table}[t]
	\caption{Comparisons among various stem modules in terms of model parameters (Param.), FLOPs, frames per second (FPS), and mIoU scores (\%).}
	\label{tab:fmvnet_stem_variants}
	\centering
	\scalebox{1.0}{
		\begin{tabular}{l|c|c|c|c}
			\hline
			Stem Modules  & Param.  & FLOPs    & FPS    & mIoU  \\ \hline \hline
			FMVNet-Stem   & 59.25M  & 1869.53G & 10.41  & 68.6  \\ \hline
			PointNet-Stem & 59.26M  & 1870.70G & 10.37  & 67.6  \\ \hline
			CENet-Stem    & 59.33M  & 1879.34G & 10.28  & 68.2  \\ \hline
			MM-Stem-A     & 59.34M  & 1880.38G & 10.22  & 67.4  \\ \hline
			MM-Stem-B     & 59.32M  & 1878.34G & 10.23  & 67.1  \\ \hline
			MM-Stem-C     & 59.27M  & 1871.55G & 10.03  & 67.5  \\ \hline
	\end{tabular}}  
\end{table}

\textbf{PointNet-Stem.} Similar to the lower layers in PointNet~\cite{pointnet_2017}, we provided PointNet-Stem, consisting of three basic convolution modules. Each module contains a convolution with the kernel size of $1\times1$, a layer normalization, and an activation function. Besides, to keep the similar model capacity among various stem modules, the channels in the first two layers were set to 48 and 64, respectively (see PointNet-Stem in Fig.~\ref{fig:stem_modules}). 

\textbf{CENet-Stem.}
In FIDNet~\cite{fidnet_2021} and CENet~\cite{cenet_2022}, authors also designed four and three basic convolution modules as the stem module, but the kernel size was set to $3\times3$. Here, we utilized the CENet-Stem with three basic convolution modules for comparison. Moreover, we set the numbers of channels in the first two layers to 32 and 64, respectively (see CENet-Stem in Fig.~\ref{fig:stem_modules}).

\textbf{MM-Stem-A/B/C.}
Some researchers might think that the LiDAR data is multi-modal. The LiDAR data is different from the color image. The gray images from the R, G, and B channels can bee seen in the same modality. By contrast, the inputs, \textit{i.e.}, range, $x$-coordinates, $y$-coordinates, $z$-coordinates, intensity, and mask, should be seen as different modalities. The range indicates the distance from the target to the LiDAR sensor, but the intensity is associated with the object's reflectance and other characteristics. Taking this into consideration, we used the depthwise convolution with the kernel size of $3\times3$ in the first basic convolution module and raised the dimension to 48 so as to compensate for the loss of model capacity (see MM-Stem-A in Fig.~\ref{fig:stem_modules}). 

Besides, we provided its variants, namely MM-Stem-B and MM-Stem-C (see MM-Stem-B/C in Fig.~\ref{fig:stem_modules}). In MM-Stem-B, we also used the depthwise convolution in the second basic convolution module, and the dimension was set to 72. In MM-Stem-C, all convolution layers were set to the depthwise convolution. 

\textbf{Analysis of Stem Modules.} The comparison results were presented in Table~\ref{tab:fmvnet_stem_variants}. We saw that with FMVNet-Stem, the model achieves the best performance. Besides, the model with CENet-Stem obtains competitive segmentation performance. Moreover, we can safely conclude that explicitly processing the multi-modal inputs is not necessary because the inputs have been normalized to be zero-mean and unit variance. By comparing the results, we validated the effectiveness of the proposed FMVNet-Stem.

\subsubsection{Removal of Layer Normalization}\label{sec:remove_normalization}
When an image classification network is revised for the semantic image segmentation task, normalization layers are commonly appended to the ends of four stages. In this subsection, we checked whether these normalization layers should be removed in the field of point cloud segmentation (PCS). Experimental results were described in Table~\ref{tab:fmvnet_layer_normalization}. We saw that in the PCS task, we still need the normalization layer after each stage, although dropping the normalization layers after the stages 3 and 4 leads to a competitive mIoU score (68.4\%).

\begin{table}[t]
	\caption{Different settings of the layer normalization (LN) at the ends of four stages.}
	\label{tab:fmvnet_layer_normalization}
	\centering
	\scalebox{1}{
		\begin{tabular}{l|c}
			\hline
			Layer Normalization        & mIoU   \\ \hline \hline
			Keep All LN                & 68.6   \\ \hline   
			Remove LN After Stage 4              & 67.6   \\ \hline
			Remove LN After Stages [3, 4]        & 68.4   \\ \hline
			Remove LN After Stages [2, 3, 4]     & 67.4   \\ \hline
			Remove LN After Stages [1, 2, 3, 4]  & 67.7   \\ \hline			
	\end{tabular}}  
\end{table}

\subsection{More Performance Comparison}
In this subsection, we first showed comparison results among various segmentation models on SemanticKITTI~\cite{semantickitti_2019_behley}, SemanticPOSS~\cite{semanticposs_2020}, and nuScenes~\cite{nuscenes_panoptic}, datasets. Then, we provided time comparison results. 

\subsubsection{Comparison on the SemanticKITTI Test Dataset}
For the results on the SemanticKITTI test dataset, we directly utilized the pre-trained weights from ConvNeXt~\cite{convnext2022} to initialize our FMVNet and then fine-tuned FMVNet on the Cityscapes~\cite{cityscapes16} dataset for 160 epochs. Subsequently, we further fine-tuned FMVNet on both SemanticKITTI training and validation datasets for 50 epochs. Finally, we submitted the predictions to the benchmark and got the IoU and mIoU scores (\%). Note that in the post-processing step, we used NLA~\cite{fidnet_2021} with the window size of $7\times7$. Besides, no test-time augmentation (TTA) techniques are applied to our results for a fair comparison. The experimental results are shown in Table~\ref{tab:64x2048_kitti_test_results}.

\begin{table*}[t]
	\caption{Quantitative comparisons on the SemanticKITTI test set in terms of IoU and mIoU scores (\%). ``$\dagger$" indicates that \textbf{TTA} is applied to the results. Also, \textbf{NO TTA} is applied to our results.}
	\label{tab:64x2048_kitti_test_results}
	\centering
	\scalebox{0.78}{
		\begin{tabular}{l|c|l|c|c|c|c|c|c|c|c|c|c|c|c|c|c|c|c|c|c|c}
			\hline
			Models	& Years & mIoU &\rotatebox{90}{Car} &\rotatebox{90}{Bicycle} &\rotatebox{90}{Motorcycle} &\rotatebox{90}{Truck} &\rotatebox{90}{Other-vehicle} &\rotatebox{90}{Person} &\rotatebox{90}{Bicyclist} &\rotatebox{90}{Motorcyclist} &\rotatebox{90}{Road} &\rotatebox{90}{Parking} &\rotatebox{90}{Sidewalk} &\rotatebox{90}{Other-ground} &\rotatebox{90}{Building} &\rotatebox{90}{Fence} &\rotatebox{90}{Vegetation} &\rotatebox{90}{Trunk} &\rotatebox{90}{Terrain} &\rotatebox{90}{Pole} &\rotatebox{90}{Traffic-sign}  \\ \hline \hline
			
			SqueezeSeg~\cite{squeezeseg} & 2018 &30.8 &68.3 &18.1 &5.1 &4.1 &4.8 &16.5 &17.3 &1.2 &84.9 &28.4 &54.7 &4.6 &61.5 &29.2 &59.6 &25.5 &54.7 &11.2 &36.3   \\ \hline
			SqueezeSegV2~\cite{squeezesegv2} & 2019 &39.7 &81.8 &18.5 &17.9 &13.4 &14.0 &20.1 &25.1 &3.9 &88.6 &45.8 &67.6 &17.7 &73.7 &41.1 &71.8 &35.8 &60.2 &20.2 &36.3  \\ \hline
			
			RangeNet21~\cite{rangenet++}& 2019 &47.4 &85.4 &26.2 &26.5 &18.6 &15.6 &31.8 &33.6 &4.0 &91.4 &57.0 &74.0 &26.4 &81.9 &52.3 &77.6 &48.4 &63.6 &36.0 &50.0 \\ \hline 
			
			RangeNet53++~\cite{rangenet++}& 2019 &52.2 &91.4 &25.7 &34.4 &25.7 &23.0 &38.3 &38.8 &4.8 &91.8 &65.0 &75.2 &27.8 &87.4 &58.6 &80.5 &55.1 &64.6 &47.9 &55.9 \\ \hline
			
			SqSegV3-21~\cite{squeezesegv3_2020}& 2020 &51.6 &89.4 &33.7 &34.9 &11.3 &21.5 &42.6 &44.9 &21.2 &90.8 &54.1 &73.3 &23.2 &84.8 &53.6 &80.2 &53.3 &64.5 &46.4 &57.6 \\ \hline
			SqSegV3-53~\cite{squeezesegv3_2020}& 2020 &55.9 &92.5 &38.7 &36.5 &29.6 &33.0 &45.6 &46.2 &20.1 &91.7 &63.4 &74.8 &26.4 &89.0 &59.4 &82.0 &58.7 &65.4 &49.6 &58.9 \\ \hline
			
			FIDNet~\cite{fidnet_2021}& 2021 &59.5 &93.9 &54.7 &48.9 &27.6 &23.9 &62.3 &59.8 &23.7 &90.6 &59.1 &75.8 &26.7 &88.9 &60.5 &84.5 &64.4 &69.0 &53.3 &62.8  \\ \hline
			
			CENet$\dagger$~\cite{cenet_2022}& 2022 &64.7 &91.9 &58.6 &50.3 &40.6 &42.3 &68.9 &65.9 &43.5 &90.3 &60.9 &75.1 &31.5 &91.0 &66.2 &84.5 &69.7 &70.0 &61.5 &67.6 \\ \hline 
			
			RangeViT~\cite{rangevit_2023} & 2023 & 64.0 & 95.4 &55.8 &43.5 &29.8 &42.1 &63.9 &58.2 &38.1 &93.1 &70.2 &80.0 &32.5 &92.0 &69.0 &85.3 &70.6 &71.2 &60.8 &64.7 \\ \hline
			
			RangeFormer~\cite{rangeformer_2023}& 2023 &69.5 &94.7 &60.0 &69.7 &57.9 &64.1 &72.3 &72.5 &54.9 &90.3 &69.9 &74.9 &38.9 &90.2 &66.1 &84.1 &68.1 &70.0 &58.9 &63.1 \\ \hline
			RangeFormer$\dagger$~\cite{rangeformer_2023}& 2023 &\textbf{73.3} &96.7 &69.4 &73.7 &59.9 &66.2 &78.1 &75.9 &58.1 &92.4 &73.0 &78.8 &42.4 &92.3 &70.1 &86.6 &73.3 &72.8 &66.4 &66.6 \\ \hline

			FMVNet (Ours)  &2024 &68.0 &96.6 & 63.4 & 60.9 & 42.1 & 55.5 & 75.6 & 70.7 & 26.1 & 92.5 & 73.8 & 79.3 & 37.7 & 92.3 & 69.3 & 85.2 & 71.4 & 69.7 & 63.0 & 66.8 \\ \hline	
	\end{tabular}}
\end{table*}

\begin{table*}[t]
	\caption{Quantitative comparisons on the SemanticKITTI val set in terms of IoU and mIoU scores (\%). Note that \textbf{NO TTA} is applied to our results. SP: spherical projection; SU++: scan unfolding++; DSK: deskewing scans; SK: skewing scans; $K$NNI: range-dependent $K$-nearest neighbor interpolation. STR: scalable training from range view strategy~\cite{rangeformer_2023}; ``$\ast$": the model pre-trained on the Cityscapes dataset. ``-": no results.}
	\label{tab:64x2048_kitti_val_results}
	\centering
	\scalebox{0.725}{
		\begin{tabular}{l|c|l|c|c|c|c|c|c|c|c|c|c|c|c|c|c|c|c|c|c|c}
			\hline
			Models	& \rotatebox{90}{Projection} & mIoU &\rotatebox{90}{Car} &\rotatebox{90}{Bicycle} &\rotatebox{90}{Motorcycle} &\rotatebox{90}{Truck} &\rotatebox{90}{Other-vehicle} &\rotatebox{90}{Person} &\rotatebox{90}{Bicyclist} &\rotatebox{90}{Motorcyclist} &\rotatebox{90}{Road} &\rotatebox{90}{Parking} &\rotatebox{90}{Sidewalk} &\rotatebox{90}{Other-ground} &\rotatebox{90}{Building} &\rotatebox{90}{Fence} &\rotatebox{90}{Vegetation} &\rotatebox{90}{Trunk} &\rotatebox{90}{Terrain} &\rotatebox{90}{Pole} &\rotatebox{90}{Traffic-sign}  \\ \hline \hline
			
			RangeFormer$^\ast$~\cite{rangeformer_2023}& SP+DSK+STR &67.6 &95.3 &58.9 &73.4 &91.3 &68.0 &78.5 &87.5 &0.0 &95.1 &49.1 &82.1 &10.8 &89.2 &67.9 &85.7 &67.7 &70.4 &64.4 &52.0 \\ \hline \hline
			
			RangeNet53++~\cite{rangenet++}&SP+DSK&54.0&-&-&-&-&-&-&-&-&-&-&-&-&-&-&-&-&-&-&- \\ \hline
			RangeNet53++ (Ours) &SP+DSK&61.5 &93.4 &47.4 &63.9 &68.8 &51.6 &69.3 &81.4 &0.0 &94.4 &47.4 &81.5 &10.0 &87.1 &57.9 &84.1 &59.4 &68.7 &55.3 &47.7 \\ \hline 
			RangeNet53++ (Ours) &SP+SK &61.7 &93.3 &47.7 &64.7 &64.0 &50.7 &70.5 &82.0 &0.0 &94.6 &48.2 &81.4 &14.3 &87.0 &57.0 &84.3 &60.6 &69.0 &55.4 &48.0 \\ \hline 
			RangeNet53++ (Ours) &SU+++DSK&63.0&94.9&50.4 &68.6 &69.5 &51.8 &72.9 &83.3 &0.0 &95.2 &49.3 &82.6 &9.9  &87.9 &57.9 &85.3 &60.9 &71.3 &56.1 &49.5 \\ \hline 
			RangeNet53++ (Ours) &SU+++SK&63.6&95.1 &51.0 &68.5 &70.9 &50.8 &74.3 & 87.0 &0.0 &95.2 &49.8 &82.6 &5.0 &88.9 &61.2 &85.7 &63.4 &71.6 &57.4 &49.0 \\ \hline 
			RangeNet53++ (Ours) &SU+++SK+$K$NNI&64.4&95.1 &51.6 &72.7 &70.7 &50.2 &75.3 &87.3 &0.0 &95.6 &47.2 &83.0 &14.9 &89.5 &63.3 & 85.8 &64.2 &71.3 &56.9 &48.9 \\ \hline \hline
			
			FIDNet~\cite{fidnet_2021}& SP+DSK &60.4&-&-&-&-&-&-&-&-&-&-&-&-&-&-&-&-&-&-&-  \\ \hline
			FIDNet (Ours)& SP+DSK &63.8 &92.9 &51.1 &66.5 &82.7 &53.1 &77.5 &89.8 &0.2 &93.8 &37.5 &80.5 &15.9 &87.0 &52.8 &85.7 &64.1 &71.4 &59.4 &50.4  \\ \hline 
			FIDNet (Ours)& SP+SK  &64.0 &93.7 &48.7 &64.8 &77.6 &54.7 &77.8 &88.2 &1.3 &93.9 &41.7 &79.8 &16.7 &87.5 &55.4 &86.1 &65.2 &72.3 &59.9 &51.3
			\\ \hline 
			FIDNet (Ours)& SU+++DSK &65.6 &93.9 &54.2 &65.3 &85.2 &53.8 &79.8 &90.3 &0.0 &94.6 &46.2 &82.2 &20.8 &88.0 &54.1 &86.8 &65.6 &74.1 &58.9 &51.8 \\ \hline 
			FIDNet (Ours)& SU+++SK  &65.8 &93.6 &51.4 &73.2 &86.7 &57.4 &79.8 &90.6 &0.0 &94.3 &43.8 &82.4 &9.2  &89.8 &57.3 &86.7 &67.2 &73.0 &61.4 &51.9 \\ \hline 
			FIDNet (Ours)& SU+++SK+$K$NNI &66.0 &94.0 &52.4 &70.0 &76.9 &57.6 &79.3 &85.3&0.0 &95.0 &45.5 &82.5 &20.5 &90.1 &61.1 &87.1 &68.4 &73.3 &63.2 &51.4  \\ \hline \hline
			
			CENet~\cite{cenet_2022}& SP+DSK &63.7&-&-&-&-&-&-&-&-&-&-&-&-&-&-&-&-&-&-&- \\ \hline
			CENet (Ours)& SP+DSK &63.5 &92.4 &47.0 &67.6 &78.7 &60.5 &78.7 &85.0 &0.2 &94.0 &40.5 &80.8 &16.0 &86.5 &49.4 &85.3 &64.2 &70.0 &59.6 &50.6  \\ \hline 
			CENet (Ours)& SP+SK  &64.5 &92.9 &52.2 &66.2 &87.3 &58.4 &76.9 &89.5 &0.0 &94.1 &41.1 &80.7 &12.8 &87.9 &54.7 &85.3 &64.3 &70.1 &59.9 &51.5  \\ \hline 
			CENet (Ours)& SU+++DSK &64.9 &93.5 &52.1 &68.0 &78.4 &60.1 &80.6 &89.3 &0.1 &94.4 &41.2 &81.8 &15.9 &88.8 &53.5 &85.7 &66.0 &71.5 &60.9 &51.4 \\ \hline 
			CENet (Ours)& SU+++SK  &65.8 &94.1 &51.5 &69.3 &78.8 &61.3 &82.0 &91.6 &0.0 &94.6 &40.5 &81.8 &10.4 &90.2 &59.8 &87.9 &67.8 &76.1 &61.2 &51.3 \\ \hline 
			CENet (Ours)& SU+++SK+$K$NNI &66.3 &94.2 &49.4 &73.1 &87.6 &59.3 &80.4 &90.0 &0.0 &95.1 &38.8 &81.8 &16.8 &89.4 &58.1 &88.3 &68.0 &75.8 &63.1 &51.0 \\ \hline \hline
			
			FMVNet (Ours)& SP+DSK   &65.3 &94.6 &50.1 &70.3 &89.9 &57.2 &77.8 &87.3  &0.0 &94.5 &47.5 &83.1 &5.9  &88.2 &56.6 &85.9 &66.0 &71.6 &63.8 &49.3  \\ \hline 
			FMVNet (Ours)& SP+SK    &65.9 &94.4 &50.8 &74.6 &89.9 &53.9 &78.7 &89.6  &0.0 &94.9 &48.8 &83.1 &11.9 &88.2 &56.6 &86.1 &66.0 &71.4 &62.1 &51.3  \\ \hline
			FMVNet (Ours)& SU+++DSK &67.4 &95.8 &55.1 &77.5 &85.2 &61.0 &81.5 &91.5  &0.0 &95.2 &47.3 &84.6 &10.7 &90.2 &61.5 &87.2 &69.0 &73.7 &65.3 &47.9  \\ \hline  
			FMVNet (Ours)& SU+++SK  &67.5 &95.3 &51.8 &78.3 &89.7 &57.9 &80.8 &90.2  &0.0 &95.5 &49.2 &84.6 &14.4 &90.4 &61.8 &87.2 &68.7 &73.6 &64.8 &48.4  \\ \hline 
			FMVNet (Ours)& SU+++SK+$K$NNI &68.6 &96.4 &55.3 &78.7 &89.5 &62.8 &82.3 &92.1  &0.0 &95.6 &47.3 &84.7 &21.2 &90.7 &64.1 &86.8 &69.7 &72.3 &63.7 &50.6  \\ \hline 
			
			FMVNet$^\ast$ (Ours)& SU+++SK+$K$NNI &\textbf{69.0} &96.7 &56.7 &77.3 &91.1 &67.3 &84.5 &94.2 &1.1 &95.8 &49.8 &85.4 &10.4 &90.9 &60.6 &87.6 &70.6 &72.7 &65.4 &52.0  \\ \hline \hline

			Fast FMVNet (Ours) & SU+++SK+$K$NNI  &67.4&96.1 &50.3 &74.0 &88.6 &67.4 &82.2 &91.1 &0.0 &95.5 &49.2 &83.8 &9.1 &90.6 &63.0 &86.1 &70.4 &70.1 &63.8 &50.2 \\ \hline
			Fast FMVNet$^\ast$ (Ours) & SU+++SK+$K$NNI &67.9 &95.3 &52.2 &76.9 &91.6 &52.0 &80.8 &91.7 &0.1 &95.8 &60.7 &84.5 &13.9 &91.2 &65.1 &86.4 &70.2 &70.9 &61.6 &49.3 \\ \hline
	\end{tabular}}
\end{table*}

We see that without test-time augmentation techniques, FMVNet achieves a higher mIoU score than the recent work RangeViT~\cite{rangevit_2023}. Besides, compared with RangeFormer~\cite{rangeformer_2023}, FMVNet achieves a competitive result. However, Table~\ref{tab:time_models} will prove that our model can achieve a better speed-accuracy trade-off.

\subsubsection{Comparison on the SemanticKITTI Validation Dataset}\label{sec:comparison_kitti_val}
We trained all models on the training dataset for 50 epochs and reported the results on the SemanticKITTI validation dataset. For fair comparisons, we reproduced RangeNet53++~\cite{rangenet++}, FIDNet~\cite{fidnet_2021}, and CENet~\cite{cenet_2022} as the baselines with the same inputs, \textit{i.e.}, \textit{ranges}, \textit{$x$-coordinates}, \textit{$y$-coordinates}, \textit{$z$-coordinates}, \textit{remissions}, and \textit{mask}. Also, we used the same data augmentation techniques and the same learning rate during the training phase. Besides, for the post-processing methods, we used $K$NN for RangeNet53++ and NLA for both FIDNet and CENet. We set the window size of $7\times7$ to the $K$NN and NLA. Moreover, we did not apply any test-time augmentation techniques to our results. The experimental results are described in Table~\ref{tab:64x2048_kitti_val_results}.

In Table~\ref{tab:64x2048_kitti_val_results}, for the results in the row of ``RangeFormer$^\ast$~\cite{rangeformer_2023}", we copied them from the paper~\cite{rangeformer_2023}. For the results ``RangeNet53++~\cite{rangenet++}", ``FIDNet~\cite{fidnet_2021}", and ``CENet~\cite{cenet_2022}", we copied them from the paper UniSeg~\cite{openpcseg2023}. The results in the rows ``RangeNet53++ (Ours)/SP+DSK" and ``FIDNet (Ours)/SP+DSK" are better than that in UniSeg, \textit{i.e.}, 61.5\% vs. 54.0\% for RangeNet53++, and 63.8\% vs. 60.4\% for FIDNet. For CENet in the rows ``CENet~\cite{cenet_2022}/SP+DSK" and ``CENet (Ours)/SP+DSK", our result is slightly lower than that in UniSeg (\textit{i.e.}, 63.5\% vs. 63.7\%). The above results show that the reproduced baselines are reasonable, and the following comparisons are fair.

Table~\ref{tab:64x2048_kitti_val_results} shows that four models trained on ``SU+++SK+$K$NNI" based images consistently achieve higher mIoU scores than their counterparts trained on ``SP+DSK" based images. The results prove the effectiveness of the proposed scan unfolding++ and range-dependent $K$-nearest neighbor interpolation (See Sec.~\ref{sec:model_on_su++_sp} and Sec.~\ref{sec:model_on_knni}). Besides, our FMVNet achieves the best mIoU score (\textit{i.e.}, 69.0\%) when it is pre-trained on the ImageNet-1K~\cite{imagenet} and Cityscapes \cite{cityscapes16} datasets (see the ``FMVNet$^\ast$ (Ours)/SU+++SK+$K$NNI" row). In addition, our Fast FMVNet with the pre-trained weights obtains the 67.9\% mIoU score while keeping 48.10 FPS (see the last row in Table~\ref{tab:64x2048_kitti_val_results} and the last row in Table~\ref{tab:fmvnet_depth_channels}). According to the results in Tables~\ref{tab:64x2048_kitti_test_results} and~\ref{tab:64x2048_kitti_val_results}, we validated the effective designs of FMVNet and Fast FMVNet.

\subsubsection{Comparison on SemanticPOSS}
\begin{table*}[t]
	\caption{Quantitative comparisons on the SemanticPOSS test set (\textit{i.e.}, sequence $\left\{02\right\}$) in terms of IoU and mIoU scores (\%). Note that \textbf{NO TTA} is applied to our results. ``$\ast$": the model pre-trained on Cityscapes~\cite{cityscapes16}.}
	\label{tab:64x2048_poss_test_results}
	\centering
	\scalebox{0.87}{
		\begin{tabular}{l|l|c|c|c|c|c|c|c|c|c|c|c|c|c}
			\hline
			Models &mIoU &People &Rider &Car &Trunk &Plants &Traffic Sign &Pole &Trashcan &Building &Cone/Stone &Fence &Bike &Ground \\ \hline \hline
			SqueezeSeg~\cite{squeezeseg}     &18.9 &14.2 &1.0  &13.2 &10.4 &28.0 &5.1  &5.7  &2.3  &43.6 &0.2  &15.6 &31.0 &75.0 \\ \hline 
			SqueezeSegV2~\cite{squeezesegv2} &30.0 &48.0 &9.4  &48.5 &11.3 &50.1 &6.7  &6.2  &14.8 &60.4 &5.2  &22.1 &36.1 &71.3 \\ \hline
			MINet~\cite{minet_2021}          &43.2 &62.4 &12.1 &63.8 &22.3 &68.6 &16.7 &30.1 &28.9 &75.1 &28.6 &32.2 &44.9 &76.3 \\ \hline \hline
			
			RangeNet53++~\cite{rangenet++}  &30.9 &57.3 &4.6  &35.0 &14.1 &58.3 &3.9  &6.9  &24.1 &66.1 &6.6  &23.4 &28.6 &73.5 \\ \hline
			RangeNet53++ (Ours)             &51.4 &74.6 &22.6 &79.8 &26.9 &71.3 &21.3 &28.2 &31.6 &77.5 &49.3 &51.7 &54.9 &77.9 \\ \hline \hline
			
			FIDNet~\cite{fidnet_2021} &46.4 &72.2 &23.1 &72.7 &23.0 &68.0 &22.2 &28.6 &16.3 &73.1 &34.0 &40.9 &50.3 &79.1 \\ \hline
			FIDNet (Ours)             &53.5 &78.5 &29.6 &79.0 &25.8 &71.4 &23.3 &32.8 &38.4 &79.2 &49.4 &54.4 &55.9 &78.2 \\ \hline	\hline
			
			CENet~\cite{cenet_2022}  &50.3 &75.5 &22.0 &77.6 &25.3 &72.2 &18.2 &31.5 &48.1 &76.3 &27.7 &47.7 &51.4 &80.3 \\ \hline 
			CENet (Ours)             &54.3 &78.1 &29.0 &83.0 &26.4 &70.5 &22.9 &33.6 &36.6 &79.2 &58.1 &53.1 &56.2 &79.6 \\ \hline \hline
			
			Fast FMVNet (Ours)       &54.3 &78.7 &27.3 &82.6 &26.6 &73.1 &25.4 &32.4 &39.0 &81.7 &45.8 &54.9 &57.6 &80.3 \\ \hline
			Fast FMVNet$\ast$ (Ours) &54.7 &80.1 &29.2 &83.9 &26.7 &73.1 &24.8 &32.7 &40.8 &81.4 &48.8 &54.8 &56.3 &78.4 \\ \hline
			FMVNet      (Ours)       &54.4 &78.7 &30.2 &80.7 &24.5 &73.2 &26.0 &35.0 &35.6 &82.8 &53.5 &51.5 &56.6 &79.5 \\ \hline
			FMVNet$^\ast$ (Ours)     &\textbf{55.1} &80.0 &29.9 &84.2 &26.2 &73.4 &25.5 &31.4 &34.9 &82.5 &55.0 &55.9 &56.4 &80.9 \\ \hline
	\end{tabular}}
\end{table*}

\begin{table*}[htp]
	\caption{Quantitative comparisons on the nuScenes validation set in terms of IoU and mIoU scores (\%). Note that \textbf{NO TTA} is applied to our results. ``Constr. Veh.": ``Construction Vehicle"; ``Drive. Sur.": ``Driveable Surface"; STR: scalable training from range view strategy~\cite{rangeformer_2023}; $\ddagger$: the model pre-trained on ImageNet-21K~\cite{imagenet}; $\ast$: the model pre-trained on Cityscapes~\cite{cityscapes16}.}
	\label{tab:32x1088_nuscenes_val_results}
	\centering
	\scalebox{0.88}{
		\begin{tabular}{l|l|c|c|c|c|c|c|c|c|c|c|c|c|c|c|c|c}
			\hline
			Models	 &mIoU &\rotatebox{90}{Barrier} &\rotatebox{90}{Bicycle} &\rotatebox{90}{Bus} &\rotatebox{90}{Car} &\rotatebox{90}{Constr. Veh.} &\rotatebox{90}{Motorcycle} &\rotatebox{90}{Pedestrian} &\rotatebox{90}{Traffic Cone} &\rotatebox{90}{Trailer} &\rotatebox{90}{Truck} &\rotatebox{90}{Drive. Sur.} &\rotatebox{90}{Other Flat} &\rotatebox{90}{Sidewalk} &\rotatebox{90}{Terrain} &\rotatebox{90}{Manmade} &\rotatebox{90}{Vegetation}  \\ \hline \hline
			RangeNet53++~\cite{rangenet++,nuscenes_panoptic} &65.5 &66.0 &21.3 &77.2  &80.9 &30.2 &66.8 &69.6 &52.1 &54.2 &72.3 &94.1 &66.6 &63.5 &70.1 &83.1 &79.8 \\ \hline
			RangeNet53++~\cite{rangenet++,openpcseg2023}& 65.8 &-&-&-&-&-&-&-&-&-&-&-&-&-&-&-&-  \\ \hline
			RangeNet53++ (Ours)            &71.1 &58.5 &38.1 &90.0  &84.0 &46.1 &80.1 &62.3 &42.3 &62.4 &80.9 &96.5 &73.7 &75.1 &74.2 &87.6 &86.0 \\ \hline \hline
			
			FIDNet~\cite{fidnet_2021,openpcseg2023} &71.8 &-&-&-&-&-&-&-&-&-&-&-&-&-&-&-&- \\ \hline
			FIDNet (Ours)             &73.5 &59.5 &44.2 &88.4 &84.6 &48.1 &84.0 &70.4 &59.9 &65.7 &78.0 &96.5 &71.6 &74.7 &75.1 &88.7 &87.3 \\ \hline \hline
			
			CENet~\cite{cenet_2022,openpcseg2023} &73.4 &-&-&-&-&-&-&-&-&-&-&-&-&-&-&-&- \\ \hline 
			CENet (Ours)            &73.4 &60.2 &43.0 &88.0 &85.0 &53.6 &70.4 &71.0 &62.5 &65.6 &80.1 &96.6 &72.3 &74.9 &75.1 &89.1 &87.7 \\ \hline \hline
			
			RangeViT$\ddagger$~\cite{rangevit_2023} &74.8 &75.1 &39.0 &90.2 &88.4 &48.0 &79.2 &77.2 &66.4 &65.1 &76.7 &96.3 &71.1 &73.7 &73.9 &88.9 &87.1 \\ \hline 
			RangeViT$^\ast$~\cite{rangevit_2023} &75.2 &75.5 &40.7 &88.3 &90.1 &49.3 &79.3 &77.2 &66.3 &65.2 &80.0 &96.4 &71.4 &73.8 &73.8 &89.9 &87.2 \\ \hline \hline
			
			RangeFormer+STR$^\ast$~\cite{rangeformer_2023}&77.1 &76.0 &44.7 &94.2 &92.2 &54.2 &82.1 &76.7 &69.3 &61.8 &83.4 &96.7 &75.7 &75.2 &75.4 &88.8 &87.3 \\ \hline
			RangeFormer$^\ast$~\cite{rangeformer_2023}&\textbf{78.1} &78.0 &45.2 &94.0 &92.9 &58.7 &83.9 &77.9 &69.1 &63.7 &85.6 &96.7 &74.5 &75.1 &75.3 &89.1 &87.5 \\ \hline \hline 
			
			Fast FMVNet (Ours)       &75.6 &60.3 &45.4 &89.9 &86.6 &55.1 &85.3 &75.1 &64.2 &67.0 &84.6 &96.7 &72.0 &75.0 &74.5 &89.5 &87.9 \\ \hline 
			Fast FMVNet$^\ast$ (Ours)&76.0 &60.3 &45.8 &95.1 &86.7 &54.7 &85.7 &74.0 &66.2 &67.1 &83.5 &96.7 &72.7 &75.1 &74.8 &89.8 &88.3 \\ \hline 
			
			FMVNet      (Ours)       &76.7 &61.5 &50.0 &94.7 &86.9 &59.0 &87.3 &78.0 &54.4 &69.1 &85.1 &97.0 &74.1 &76.3 &75.7 &90.2 &88.7 \\ \hline 
			FMVNet$^\ast$ (Ours)     &76.8 &61.1 &49.5 &94.7 &86.8 &59.6 &71.1 &77.2 &69.1 &70.9 &85.6 &96.9 &75.0 &76.5 &75.8 &90.1 &88.4 \\ \hline 
	\end{tabular}}
\end{table*}

For fair comparisons, we trained RangeNet53++~\cite{rangenet++}, FIDNet~\cite{fidnet_2021}, and CENet~\cite{cenet_2022} with the same inputs and data augmentation techniques. Besides, we trained all our models for 50 epochs on the SemanticPOSS~\cite{semanticposs_2020} training dataset and reported results on the test dataset. Other experimental settings are the same as that in Sec.~\ref{sec:comparison_kitti_val}. The experimental results were provided in Table~\ref{tab:64x2048_poss_test_results}.

The results in the rows of ``RangeNet53++~\cite{rangenet++}", ``FIDNet~\cite{fidnet_2021}", and ``CENet~\cite{cenet_2022}" were copied from the work~\cite{cenet_2022}. We saw that the reproduced models achieve better performance than the counterparts, \textit{i.e.}, 51.4\% vs. 30.9\% for RangeNet53++, 53.5\% vs. 46.4\% for FIDNet, and 54.3 vs. 50.3 for CENet. The performance gains can be attributed to the proposed scan unfolding++ (SU++) and range-dependent $K$-nearest neighbor interpolation ($K$NNI). Besides, the proposed FMVNet achieves the 54.4\% mIoU score. With the pre-trained weights, the mIoU score of FMVNet is further increased to 55.1\%. Moreover, Fast FMVNet also obtains competitive results, \textit{i.e.}, 54.3\% and 54.7\% mIoU scores. The experimental results can validate the effectiveness of the proposed SU++, $K$NNI, FMVNet, and Fast FMVNet. 

\subsubsection{Comparison on nuScenes}
Similar to the previous experiments, the reproduced RangeNet53++~\cite{rangenet++}, FIDNet~\cite{fidnet_2021}, and CENet~\cite{cenet_2022} were trained on the nuScenes dataset with the same inputs and data augmentation techniques. Besides, we trained all models for 80 epochs on the training dataset and reported mIoU and IoU scores on the validation dataset. Other experimental settings are the same as that in Sec.~\ref{sec:comparison_kitti_val}. The experimental results were provided in Table~\ref{tab:32x1088_nuscenes_val_results}. 

In Table~\ref{tab:32x1088_nuscenes_val_results}, the results in the ``RangeNet53++~\cite{rangenet++,nuscenes_panoptic}" were copied from the work~\cite{nuscenes_panoptic}. The results in the rows of ``RangeNet53++~\cite{rangenet++,openpcseg2023}", ``FIDNet~\cite{fidnet_2021,openpcseg2023}", and ``CENet~\cite{cenet_2022,openpcseg2023}" were copied from the paper UniSeg~\cite{openpcseg2023}. In Table~\ref{tab:32x1088_nuscenes_val_results}, the reproduced RangeNet53++ and FIDNet achieve better performance than their counterparts. For CENet, we obtained the same result as that in UniSeg. The performance gains of the reproduced RangeNet53++, FIDNet, and CENet can validate the effectiveness of the proposed scan unfolding++ and range-dependent $K$-nearest neighbor interpolation. Besides, the proposed FMVNet and Fast FMVNet get 76.7\% and 75.6\% mIoU scores, respectively. Moreover, after pre-trained on the Cityscapes dataset, FMVNet and Fast FMVNet obtain 76.0\% and 76.8\% mIoU scores. Furthermore, Table~\ref{tab:32x1088_nuscenes_val_results} shows that FMVNet is inferior to RangeFormer. However, after checking the nuScenes~\cite{nuscenes_panoptic} validation dataset, we found that at least 5.7\% of total points are erroneously labelled. Specifically, all points with $\boldsymbol{x} \notin \left[-50m, 50m\right]$, $\boldsymbol{y} \notin \left[-50m, 50m\right]$, and $\boldsymbol{z} \notin \left[-5m, 3m\right]$ should be annotated as ``ignored"~\cite{lasermix2023}. More importantly, according to these constraints, we removed these points during the training phase. Hence, our FMVNet and Fast FMVNet only achieve suboptimal performance.

\subsubsection{Qualitative Comparisons on SemanticKITTI}

\begin{figure*}[t]
	\centering
	\includegraphics[width=1.97\columnwidth]{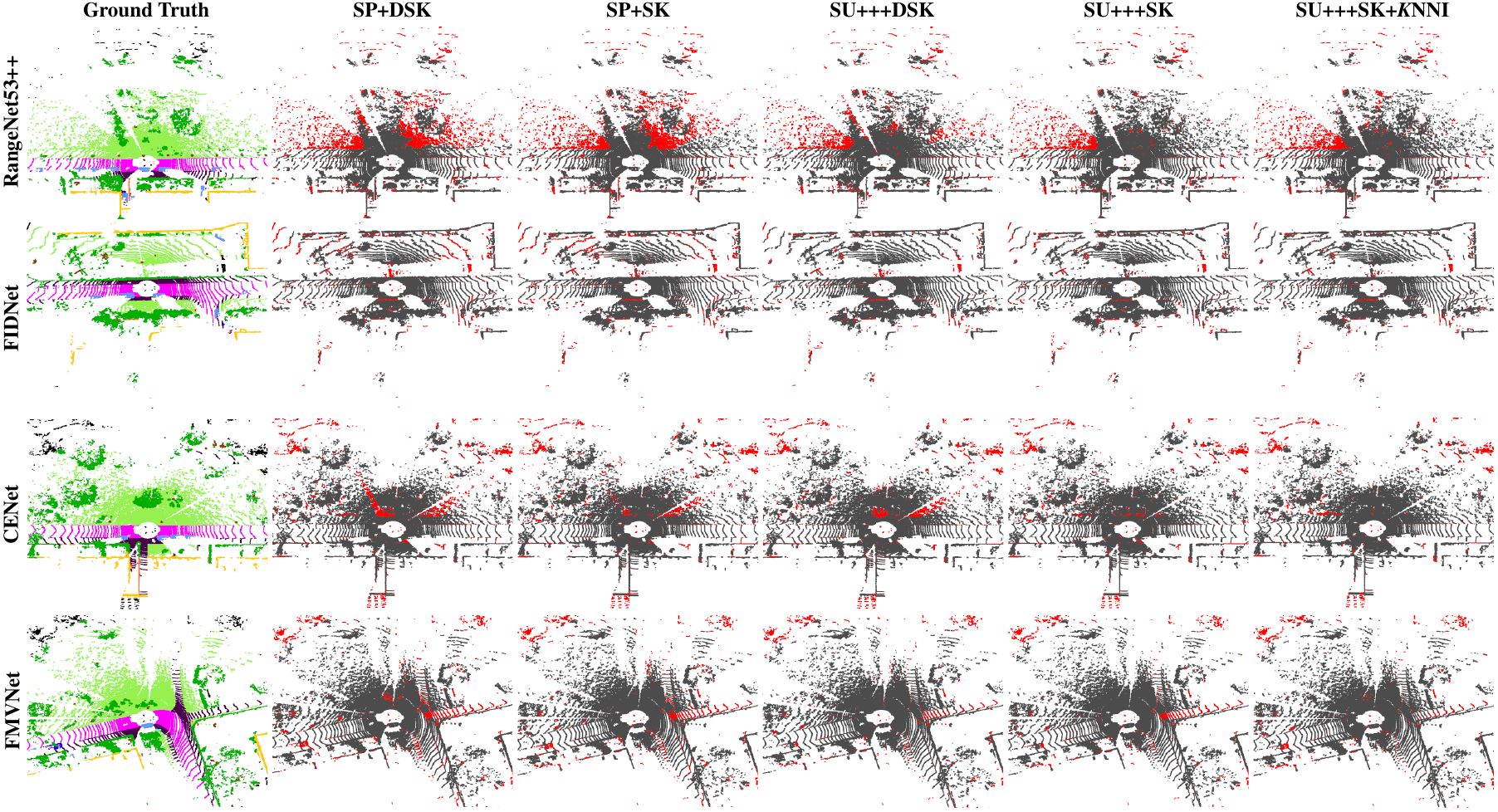}
	\caption{Qualitative comparisons among RangeNet53++, FIDNet, CENet, and our FMVNet trained on ``SP+DSK", ``SP+SK", ``SU+++DSK", ``SU+++SK", ``SU+++SK+$K$NNI" based images. Correct and incorrect predictions are indicated by gray and red colors, respectively. ``SP": spherical projection; ``DSK": deskewing scans; ``SK": skewing scans; ``SU++": scan unfolding++; ``$K$NNI": range-dependent $K$-nearest neighbor interpolation.}
	\label{fig:models_data_comparisons}
\end{figure*}

We here provided qualitative comparisons of RangeNet53++~\cite{rangenet++}, FIDNet~\cite{fidnet_2021}, CENet~\cite{cenet_2022}, and our FMVNet which are trained on ``SP+DSK", ``SP+SK", ``SU+++DSK", ``SU+++SK", and ``SU+++SK+$K$NNI" based images, respectively. The experiments were conducted on the SemanticKITTI~\cite{semantickitti_2019_behley} validation set (see Fig.~\ref{fig:models_data_comparisons}).

Fig.~\ref{fig:models_data_comparisons} shows that the segmentation models trained on the ``SU+++SK+$K$NNI" range images are able to accurately segment the point cloud (see the last column in Fig.~\ref{fig:models_data_comparisons}). This suggests that range image-based segmentation models can benefit from the images with coherent and complete objects generated by the proposed SU++ and $K$NNI. 

\begin{table}[t]
	\caption{Time comparisons among spherical projection (SP), scan unfolding++, and range-dependent $K$-nearest neighbor interpolation ($K$NNI) under various range image sizes on the SemanticKITTI validation set. Average time on each scan is reported (Unit: millisecond (ms))}
	\label{tab:time_consumed}
	\centering
	\scalebox{1.0}{
		\begin{tabular}{l|c|c|c}
			\hline
			Methods     &$64\times512$ &$64\times1024$ &$64\times2048$  \\ \hline \hline        
			SP          & 15.76ms & 16.32ms  & 17.32ms   \\ \hline
			SU++        & 14.68ms & 15.30ms  & 16.28ms   \\ \hline
			SU+++$K$NNI & 15.20ms & 16.18ms  & 18.89ms   \\ \hline               
	\end{tabular}}  
\end{table}

\begin{table}[t]
	\caption{Comparisons among the models in terms of the number of model parameters (Params.), latency, frames per second (FPS), and mIoU scores (\%) on the SemanticKITTI~\cite{semantickitti_2019_behley} validation dataset. ``$\ast$": our reproduced models; ``-BN": FMVNet with batch normalization.}
	\label{tab:time_models}
	\centering
	\scalebox{0.95}{
		\begin{tabular}{l|c|c|c|c|c}
			\hline
			Methods                              & Years &Params. & Latency &FPS   & mIoU  \\ \hline  \hline
			MinkowskiNet~\cite{minkowski2019}    & 2019  &21.7M   & 48.4ms  &20.7  & 61.1  \\ \hline  
			RangeNet53++$^\ast$~\cite{rangenet++}& 2019  &50.4M   & 13.9ms  &71.9  & 64.4  \\ \hline
			Cylinder3D~\cite{cylindrical3d2021}  & 2021  &56.3M   & 71.5ms  &13.3  & 65.9  \\ \hline  
			FIDNet$^\ast$~\cite{fidnet_2021}     & 2021  &6.1M    & 16.2ms  &61.8  & 66.0  \\ \hline
			CENet$^\ast$~\cite{cenet_2022}       & 2022  &6.8M    & 15.5ms  &64.5  & 66.3  \\ \hline
			RangeFormer~\cite{rangeformer_2023}  & 2023  &24.3M   & 90.3ms  &11.1  & 67.6  \\ \hline
			UniSeg 0.2$\times$~\cite{openpcseg2023}& 2023&28.8M   & 84.6ms  &11.8  & 67.0  \\ \hline 
			UniSeg 1.0$\times$~\cite{openpcseg2023}& 2023&147.6M  & 145.0ms &6.9   & 71.3  \\ \hline \hline
			Fast FMVNet  (Ours)                  & 2024  &4.3M    & 20.8ms  &48.1  & 67.9  \\ \hline
			FMVNet-BN (Ours)                     & 2024  &59.3M   & 64.8ms  &15.4  & 68.3  \\ \hline
			FMVNet (Ours)                        & 2024  &59.3M   & 96.1ms  &10.4  & 69.0  \\ \hline 
	\end{tabular}}  
\end{table}

\begin{figure}[t]
	\centering
	\includegraphics[width=0.75\columnwidth]{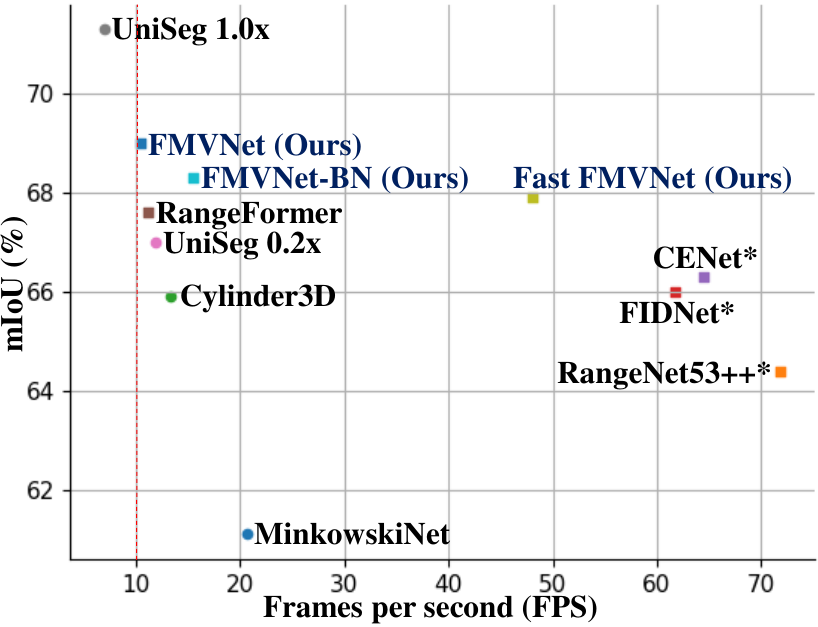}
	\caption{Comparison results among various models in terms of frames per second (FPS) and mIoU scores (\%). The marker ``$\square$" indicates range image-based models. ``$\ast$": reproduced models.}
	\label{fig:speed_miou}
\end{figure}

\subsubsection{Time Comparisons}
Time comparison results about the pre-processing step and the models were provided here.

\textbf{The Pre-processing Step.} The computational cost in the pre-processing step is an important factor to consider, especially in robotic applications. We here drew comparisons among spherical projection (SP), scan unfolding++ (SU++), and range-dependent $K$-nearest neighbor interpolation ($K$NNI) in terms of running time. Specifically, we adopted SP and SU++ to generate all range images and corresponding look-up tables on the SemanticKITTI validation set (\textit{i.e.}, sequence $\{08\}$). For $K$NNI, we used SU++ and $K$NNI together because the consumption time of $K$NNI is limited. Then, the average running time on each scan was utilized to compare these methods. The experiments were conducted on a desktop computer with a CPU ``Intel Core i9-10900K @3.70GHz" and a ``DDR4 RAM 32GB (16GB$\times$2)". The comparison results under various sizes of range images were provided in Table~\ref{tab:time_consumed}.

In Table~\ref{tab:time_consumed}, we saw that among SP, SU++, and SU+++$K$NNI, SU++ spends the least time in producing the range image with various sizes, because SU++ does not need to compute the vertical coordinates. Moreover, under the sizes of $64\times512$ and $64\times1024$, the times spent by SU+++$K$NNI are less than that by SP. This is because there are not many missing points to fill in. By contrast, when the range image size is set to $64\times2048$, SU+++$K$NNI takes the most time to process one range image. The experimental results validate the efficiency of the proposed SU++ and $K$NNI.

\textbf{The Models.} Comparison results in terms of efficiency and mIoU scores (\%) among various models were provided in Table~\ref{tab:time_models} and Fig.~\ref{fig:speed_miou}. For fair comparisons, we used the best mIoU scores (\%) of RangeNet53++, FIDNet, and CENet on the SemanticKITTI validation dataset. The results of MinkowskiNet~\cite{minkowski2019}, Cylinder3D~\cite{cylindrical3d2021}, UniSeg 0.2$\times$~\cite{openpcseg2023}, and UniSeg 1.0$\times$ were copied from the UniSeg paper. Moreover, we reproduced RangeFormer~\cite{rangeformer_2023} because no open-source code was found. The architecture of RangeFormer is very similar to that of SegFormer-B2~\cite{segformer2021}, so we modified SegFormer-B2 towards RangeFormer based on the description in the paper. Also, we set the input image size to $6\times64\times1920$ according to the STR~\cite{rangeformer_2023}, but we did not run RangeFormer five times or stack five sub point clouds in the single forward pass. Additionally, the size of inputs to all our models was set to $6\times64\times2048$. All models were tested on the NVIDIA A100 GPU.

Table~\ref{tab:time_models} shows that Fast FMVNet achieves a higher mIoU score than UniSeg 0.2$\times$ and RangeFormer, and is about four times faster than the two models. Besides, compared with RangeNet53++, FIDNet, and CENet, Fast FMVNet obtains the best performance and has fewer model parameters (\textit{i.e.}, only 4.3M). Moreover, with batch normalization instead of layer normalization, our FMVNet can get 15.4 FPS and still achieve competitive performance (\textit{i.e.}, 68.3\% mIoU score). Besides, all our models can meet the speed requirement, \textit{i.e.}, executing at least 10 scans per second (see Fig.~\ref{fig:speed_miou}). The high execution speed of FMVNet and Fast FMVNet is attributed to the range image-based input and convolution-based network architecture. The comparison results show that Fast FMVNet achieves a better speed-accuracy trade-off than other models.

\section{Discussions}\label{sec:discussions}
In this section, We discuss the LiDAR data, interpolation method, test-time augmentation, limitations, and potential impact. 

\subsection{What kind of LiDAR data is suitable for range image-based point cloud segmentation?} Raw LiDAR data without motion compensation is preferable. This can avoid the massive missing points along the horizontal direction when they are projected onto the range image. Note that all point clouds in SemanticKITTI~\cite{semantickitti_2019_behley} have been calibrated. 

Besides, for each point, additional values such as the \textit{laser id (or ring number)}, \textit{azimuth angle}, and \textit{vertical angle} should be provided. Ring numbers can be used to unfold the point cloud and help avoid the massive point overlapping in the vertical direction. This is why we use scan unfolding++ to prepare range images in this paper. The azimuth and vertical angles are useful in augmenting input data during training. Moreover, if the point clouds include 0-distance values and outliers, the azimuth and vertical angles are useful for keeping the data structure. For example, there are many 0-distance values and a few outliers (the distances exceeding 1000 meters) in nuScenes~\cite{nuscenes_panoptic}. Without the azimuth angles for these 0-distance values, we do not know where they are in the range image. Moreover, the azimuth and vertical angles are very important for developing pointwise operations such as 1D convolution on the point clouds. Note that modern LiDAR sensors can easily output the laser id, azimuth angle, and vertical angle.


\subsection{Do we really need an interpolation method for range image-based point cloud segmentation?}
The experimental results in this paper have validated that an interpolation approach for LiDAR data is necessary. Actually, in the commonly used depth cameras such as \textit{Intel RealSense}, the corresponding software has contained the interpolation algorithms (see 
Holes Filling Filter in the document\footnote{https://dev.intelrealsense.com/docs/post-processing-filters}).

\subsection{Why do not we use test-time augmentation (TTA) techniques to achieve high IoU and mIoU scores?}
Using TTA techniques and an ensemble on the test data leads to prohibitive inference time. It is not practical in applications. Besides, utilizing these tricks can boost the performance of segmentation models but might lead to misleading results. Moreover, we expect that our models can serve as the baselines for the following range image-based approaches in the point cloud segmentation task. Therefore, we did not apply any TTA techniques and the ensemble to our results.

\subsection{What are the limitations of this work?}
The limitations are summarized as follows: (1) The proposed scan unfolding++ on the SemanticKITTI dataset can not automatically generate range images. Users need to produce the laser id for each point, skew the scans, and save the processed data before preparing the images. However, note that modern LiDAR sensors can directly produce raw LiDAR data with the laser indices (or ring numbers). Therefore, we do not need to manually make the laser indices and skew the scans in practical applications. (2) We did not adopt the grid search or other methods to tune hyper-parameters in the loss function and the learning rate. Actually, we set the same learning rate for all reproduced and proposed models during the training phase. This might result in suboptimal performance for the models. Choosing an optimal set of hyper-parameters can boost the performance of models.

\subsection{What is the potential impact on the community?} 
The proposed SU++ and $K$NNI can be employed for other range image-based tasks, such as moving object segmentation~\cite{movingobjseg2021}, simulation-to-real domain adaptation~\cite{epointda2021,unsuperlida2022,adversar2023}, large-scale point cloud registration~\cite{regformer2023}, and simultaneous localization and mapping~\cite{semantic_slam_2019,overlaptrans2022}. Besides, the proposed methods might be beneficial to multimodal models trained on both natural images and range images.

\section{CONCLUSION}\label{sec:conclusion}
Point cloud segmentation plays a crucial role in robot perception and navigation tasks. In this paper, we pointed out the sources of missing values in the range images, \textit{i.e.}, the unreasonable projection approach, the deskewing scans, and the inherent properties of the LiDAR sensor. The missing values in the images decrease the performance of segmentation models by damaging the shapes and patterns of objects. To fill in missing values, we proposed scan unfolding++ (SU++) to generate range images. Furthermore, we proposed an embarrassingly simple range-dependent $K$-nearest neighbor interpolation ($K$NNI) to fill in undesirable missing values further. Besides, we introduced the Filling Missing Values Network (FMVNet) and Fast FMVNet to achieve state-of-the-art performance in terms of efficiency and accuracy. The experimental results on the SemanticKITTI, SemanticPOSS, and nuScenes datasets demonstrated that the segmentation models trained on the ``SU+++SK+$K$NNI" based range images consistently achieve better performance than their counterparts trained on the ``SP+DSK" based images. This validates the effectiveness of the proposed SU++ and $K$NNI. Besides, our FMVNet can execute more than 10 FPS and achieve competitive performance. Our Fast FMVNet can achieve a better speed-accuracy trade-off than existing models.






\bibliographystyle{IEEEtran}
\bibliography{icra2024.bib}

\end{document}